\documentclass[conference]{IEEETran}
\usepackage[utf8]{inputenc}
\usepackage[T1]{fontenc}    
\usepackage{hyperref}       
\usepackage{url}            
\usepackage{booktabs}       
\usepackage{amsfonts}       
\usepackage{nicefrac}       
\usepackage{microtype}      
\usepackage{xcolor}         
\usepackage{amsmath,amssymb}
\usepackage{wrapfig}
\usepackage{stfloats}
\usepackage{graphicx}
\usepackage{float}
\usepackage{caption}
\usepackage{subcaption}
\usepackage{bm}
\usepackage{listings}
\usepackage{multicol}
\usepackage{multirow}

\NewDocumentCommand{\codeword}{v}{%
\texttt{\textcolor{blue}{#1}}%
}

\title{A \textbf{T}emporal \textbf{L}inear \textbf{N}etwork for Time Series Forecasting}

\author{%
  \IEEEauthorblockN{%
    Rémi Genet\IEEEauthorrefmark{1}\textsuperscript{\textsection} and
    Hugo Inzirillo\IEEEauthorrefmark{2}\textsuperscript{\textsection}
  }%
  \IEEEauthorblockA{\IEEEauthorrefmark{1} DRM, Université Paris Dauphine - PSL}%
  \IEEEauthorblockA{\IEEEauthorrefmark{2} CREST-ENSAE, Institut Polytechnique de Paris}%
}

\begin{document}

\thispagestyle{plain}
\pagestyle{plain}
\maketitle
\begingroup\renewcommand\thefootnote{\textsection}
\footnotetext{These authors contributed equally.}
\endgroup

\begin{abstract}
Recent research has challenged the necessity of complex deep learning architectures for time series forecasting, demonstrating that simple linear models can often outperform sophisticated approaches. Building upon this insight, we introduce a novel architecture the Temporal Linear Net (TLN), that extends the capabilities of linear models while maintaining interpretability and computational efficiency. TLN is designed to effectively capture both temporal and feature-wise dependencies in multivariate time series data. Our approach is a variant of TSMixer that maintains strict linearity throughout its architecture. TSMixer removes activation functions, introduces specialized kernel initializations, and incorporates dilated convolutions to handle various time scales, while preserving the linear nature of the model. Unlike transformer-based models that may lose temporal information due to their permutation-invariant nature, TLN explicitly preserves and leverages the temporal structure of the input data. A key innovation of TLN is its ability to compute an equivalent linear model, offering a level of interpretability not found in more complex architectures such as TSMixer. This feature allows for seamless conversion between the full TLN model and its linear equivalent, facilitating both training flexibility and inference optimization. Importantly, we demonstrate that TLN outperforms standard linear regression models. This superior performance is attributed to the unique training structure of the TLN and the inherent relationship of weights established during model construction, which can not be found in conventional regression approaches. Our findings suggest that TLN strikes a balance between the simplicity of linear models and the expressiveness needed for complex time series forecasting tasks. It offers improved interpretability compared to TSMixer while demonstrating greater resilience in multivariate cases than basic linear models and standard linear regression. This work contributes to the ongoing discussion about the trade-offs between model complexity and performance in time series analysis, and opens new possibilites for developing efficient and interpretable forecasting models that leverage the strengths of neural network architectures and linear models.
\end{abstract}

\section{Introduction}
Time series forecasting is a critical task with wide-ranging applications in various domains such;  traffic flow estimation, energy management, and financial investment \cite{zeng2023transformers}. 
Multivariate time series have attracted a great interest in the scientific community \cite{wei2018multivariate,marcellino2006comparison}, becoming an essential field of study for many industrial applications. The constant increase in available data justifies the development of increasingly sophisticated predictive frameworks. Unlike univariate time series that focus on the evolution of a single variable, multivariate series framework integrate several interdependent variables which significantly complicates the learning task. The analysis of these data requires models capable of managing temporal dependencies and dynamic interactions across different dimensions and time horizons. To address these challenges, researchers have explored a variety of approaches, from neural networks \cite{tang1991time} to state-space models \cite{hamilton1994state,kim1994dynamic}, vector autoregressive models \cite{zivot2006vector,lutkepohl2013vector} and advanced neural architectures \cite{binkowski2018autoregressive,ma2019novel}. Despite the increase in the availability of dataset for time series forecasting, it reamains a very complex task. Powerful models have been proposed through the years \cite{salinas2020deepar,oreshkin2019n}.  Some research proposed dynamic system to model different existing states within a time series \cite{rangapuram2018deep,li2021learning,inzirillo2024deep}. Deep learning techniques, particularly Long Short-Term Memory (LSTM) \cite{hochreiter1997long}  networks and attention \cite{vaswani2017attention} mechanisms, have shown promising results due to their ability to handle nonlinearities and long-term dependencies in data. In this paper, we introduce a novel architecture that combines the interpretability of linear models with the structural advantages of deep learning approaches. We demonstrate TLN's effectiveness and consistency through extensive experiments on two different multistep forecasting tasks, showing robust performance across varying sequence lengths and prediction horizons. Our results show that TLN achieves superior or comparable performance to more complex architectures while maintaining complete interpretability through its linear equivalent form and demonstrating remarkable stability across different input configurations. The implementation of TLN is available as an open-source Python package \texttt{temporal\_linear\_network} on PyPI, and the complete source code can be found at \href{https://github.com/remigenet/TLN/}{TLN}. This repository includes all the code necessary to reproduce our experiments and implement TLN in various time series forecasting applications.
\section{Related Work}
Econometrics methods for predicting time series, such as ARIMA \cite{box1976time} and its variants, have been the predominant choices for stationary and linear time series. These models rely on assumptions of stationarity and linearity which limit their ability to capture complex dynamics. These complex dynamics are often observed in real world data. Approaches such as Prophet \cite{taylor2018forecasting} have been proposed to improve the use of statistical models by incorporating seasonal and holiday effects, but remain limited in contexts where non-linear interactions are predominant. Over the past decades, time series forecasting solutions have evolved from traditional statistical methods to sophisticated deep learning  approaches. RNNs \cite{rumelhart1986learning}  has emerged as a fundamental architecture with the capacity of maintaining internal memory which allowed sequential data management. Limitation of standards RNNs lies in vanishing gradient \cite{bengio1994learning} which motivates the development of more sophisticated architecture, LSTMs \cite{hochreiter1997long}. LSTMs uses gating mechanism to increase the long term memory management. \cite{cho2014learning} proposed the GRU (Gated Recurrent Units) offering an alternative to LSTMs. They are less complex while maintaning comparable performances. Recent research has turned to Mixture of Experts (MoE) \cite{jordan1994hierarchical}, architectures that improve flexibility and performance by dynamically selecting specialized models or "experts" based on the input data. Many experts choice methods have been proposed \cite{masoudnia2014mixture,yuksel2012twenty} . In our previous work we proposed a method using gated mechanism relying on kolmogorov-arnold networks \cite{liu2024kan} to weight the contribution of each expert to the global output \cite{inzirillo2024gated}. We refer the reader to \cite{masoudnia2014mixture} for further information. Recently, there has been a surge in Transformer-based architectures \cite{lin2022survey} with extensions for time series analysis and forecasting \cite{lim2021temporal,genet2024temporal}, particularly for the challenging task of long-term time series forecasting (LTSF) \cite{zeng2023transformers}. Transformer have demonstrated remarkable success in natural language processing \cite{wolf2020transformers}, machine translation \cite{wang2019learning} and computer vision tasks. We mentionned it before, they have been adapted for time series forecasting, Lim et al. \cite{lim2021temporal} proposed a Temporal fusion Transformer (TFT), built using LSTMs and attention mechansim. However, the effectiveness of Transformer-based models for time series forecasting has been called into question. Zeng et al. \cite{zeng2023transformers} argue that while Transformers excel at extracting semantic correlations among elements in a long sequence, time series modeling primarily involves extracting temporal relations. The permutation-invariant nature of the self-attention mechanism in Transformers may result in temporal information loss, which is crucial for time series analysis. In contrast, Chen et al. \cite{chen2023tsmixer} propose TSMixer, a novel architecture designed by stacking multi-layer perceptrons (MLPs) \cite{hornik1989multilayer}. TSMixer is based on mixing operations along both the time and feature dimensions to extract information efficiently. This approach aims to leverage the strengths of MLPs while addressing the limitations of Transformer-based models in capturing temporal dependencies. The debate surrounding the efficacy of Transformer-based models versus simpler architectures like linear models or MLPs for time series forecasting highlights the need for further research and comparison of different approaches. This paper aims to explore and compare various models for time series forecasting, focusing on their ability to capture temporal relations and their performance on benchmark datasets.

\section{Architecture}
We introduce the Temporal Linear Network (TLN) , a novel architecture designed to capture complex temporal relationships while maintaining interpretability through its ability to be transformed into an equivalent linear model. The TLN model consists of multiple layers, each comprising a SequentialDense operation followed by a 1D convolution. For a model with $L$ layers, the forward pass can be expressed as:
\begin{equation}
    \Tilde{X}^{(l)} = f_{\omega_l}(X^{(l)}),
\end{equation}
For each feature we compute the 1D convolution along the sequence:
\begin{equation}
    \text{Conv1D}_{w_{l,q}} (\Tilde{X_q}^{(l)})) =\sum_{k=0}^{K^{(l)}-1} W^{(l)}_{k,q} \cdot  \Tilde{X}^{(l)}_{t+k, q} + b_q,
\end{equation}
where $\Tilde{X_q}^{(l)}$ is the input sequence, $W \in \mathbb{R}^{K\times Q}$ is the associated weights and $b \in \mathbb{R}^{Q}$ the bias. Writing the  equation for each layer of network, we obtain:
\begin{equation}
    \Tilde{X}^{(l)} = \text{Conv1D}_{\omega_l}(f_{\omega_l}(\Tilde{X}^{(l-1)})),
\end{equation}
where $\omega_l$ denotes all the weights shared for each layer. The final layer form will be:
\begin{equation}
    \Tilde{X}^{(L)} = \text{Conv1D}_{\omega_L}(f_{\omega_L}(\Tilde{X}^{(L-1)})).
\end{equation}

\subsection{SequentialDense Layer}
The SequentialDense layer applies separate transformations along the feature and time dimensions. For an input tensor $X \in \mathbb{R}^{S \times F}$, where $B$ denotes the batch size, $S$  the input sequence length, and $F$ is the number of input features. The SequentialDense layer performs the following operations. The first step consist in an element wise scaling:
\begin{equation}
    \Tilde{x} = K \odot X
\end{equation}
where the base kernel is represented by $K \in \mathbb{R}^{ S \times 1}$. Followed by the feature transformation
\begin{equation}
    x_{out} = F \odot (W_{x} \Tilde{x} + b_{x}).
\end{equation}
 The feature kernel is given by $F \in \mathbb{R}^{ 1 \times F'} $, where $F'$ denotes the number of output features. The final step of the sequential dense layer consist in applying a time kernel along each sequence.
\begin{equation}
    Y = \tau \odot (W_t x_{out} + B_t)^{'},
\end{equation}
 where the time kernel is $\tau \in \mathbb{R}^{S' \times 1}$, where $S'$ is the output sequence length.
 
\subsection{Comparison with Standard Linear Regression}
In a standard linear regression on flattened inputs, the model would have a single weight matrix $W \in \mathbb{R}^{(S \times F) \times (S' \times F')}$ and a bias vector $b \in \mathbb{R}^{S' \times F'}$. The prediction would be computed as:
\begin{equation}
    Y_{ijk} = \sum_{l=1}^{SF} W_{ij,l} X_{l} + b_{ij}
\end{equation}
It's important to note that in many cases, the number of parameters in this standard linear regression model $(S \times F \times S' \times F' + S' \times F')$ can be significantly larger than the number of parameters in our TLN model. This is because TLN uses parameter sharing and structured transformations to reduce the total number of parameters while still capturing complex temporal relationships.

\subsection{Equivalent Linear Model}

Despite its complex structure, TLN can be transformed into an equivalent linear model. This is possible because all operations in TLN (SequentialDense and Conv1D) are linear transformations. The composition of linear transformations is itself a linear transformation, which guarantees the existence of an equivalent linear model. Let $f_1, f_2, ..., f_n$ be linear transformations. Then, for any vectors $x$ and $y$, and any scalar $c$:
\begin{equation}
    \begin{split}
        f_n(...f_2(f_1(x + cy))) &= f_n(...f_2(f_1(x) + cf_1(y)))\\
        &= f_n(...(f_2(f_1(x)) + cf_2(f_1(y))))\\
        &= f_n(...f_2(f_1(x))) + cf_n(...f_2(f_1(y)))
    \end{split}
\end{equation}
This proves that the composition $f_n \circ ... \circ f_2 \circ f_1$ is also a linear transformation.To find the equivalent linear model, we can compute the equivalent weights $W_{eq} \in \mathbb{R}^{S \times F \times S' \times F'}$ and bias $b_{eq} \in \mathbb{R}^{S' \times F'}$ as follows:
First, compute the bias by passing a zero input through the model:
\begin{equation}
    \begin{split}
        b_{eq} &= \text{TLN}(0)\\
        &= TLN(0_{S' \times F'})
    \end{split}
\end{equation}
Then, for each input position $(i, j)$, compute:
\begin{equation}
   W_{eq}^{(i, j)} = \text{TLN}(e_{ij}) - b_{eq}
\end{equation}
where $e_{ij}$ is a unit vector with a 1 at position $(i, j)$ and 0 elsewhere. The equivalent linear model can then make predictions using:
\begin{equation}
    Y = \sum_{i=1}^{S} \sum_{j=1}^{F} W_{eq}^{(i, j)} \cdot X_{i,j} + b_{eq}
\end{equation}
This transformation allows us to interpret the complex TLN model as a single linear operation, providing insights into how the model uses different parts of the input sequence to make predictions, while maintaining a more parameter-efficient structure during training and standard inference.
\section{Learning}
To evaluate the effectiveness of the Temporal Linear Network (\textit{TLN}), we conducted experiments on two different tasks, designed to study multiple aspects of the model's performance.
\subsection{Task 1: Market Traded Notional Predictions}
Our first task focuses on predicting market traded notional values, inspired by recent works in financial time series forecasting \cite{genet2024tkan,inzirillo2024sigkan,genet2024temporal,inzirillo2024gated}. Financial markets are known for their high volatility and noise, making forecasting challenging. However, traded volume series often exhibit stronger patterns, including seasonality, autocorrelation, and correlations between different assets. For this task, we applied a min-max scaler to preprocess the data, ensuring all values fall within a consistent range.
\subsection{Task 2: ETTh1 Dataset}
For our second task, we utilized the ETTh1 dataset, a well-known benchmark in time series forecasting literature. This dataset consists of hourly measurements related to electricity transformer temperatures. To align with previous studies and enable fair comparisons, we preprocessed this data using a standard scaler.
\subsection{Experimental Setup}
For both tasks, we designed our experiments to study the impact of various factors on model performance; we varied the input sequence length by testing values of 3, 10, 20, 50, 100, 200, and 400 time steps to understand how sequence length impacts model accuracy. We also compared the performance of models using univariate inputs (single variable) versus those utilizing multivariate inputs (multiple variables) to determine the benefits of incorporating additional features. We also explored the effect of adding temporal information, such the hour and day of the week. We stacked this information to the input data  to capture any underlying patterns related to "time". Finally, we evaluated the models across different forecasting horizons. We made predictions for 1, 6, 15, 30, 90, 180, and 360 hours ahead. The objective was to assess the capability of the different models in making both short-term and long-term forecasts. Both tasks use hourly data.This configuration allows us to perform a consistent comparison across different forecasting horizons and input sequence lengths.
\subsection{Comparison}
We compared the \textit{TLN} against a diverse set of baseline models.

\medskip

\textit{Linear Regression}: This model flattens the sequential dimension, transforming the input from shape (sequence\_length, features) to (sequence\_length * features). It then applies a standard linear regression to predict the target values.

\medskip

\textit{Linear Models from "Are Transformers Effective for Time Series Forecasting?"}:
These models were a key inspiration for our work, as they demonstrate the effectiveness of simple linear approaches. While similar to TLN in their linear nature, they lack some of the mechanisms present in our model.
In our experiment, we implemented three types of linear layers to process the time series data. The first is "CLinear", which is a classic linear layer applied directly across the time dimension. This setup provides a straightforward transformation of the input data. The second, "NLinear", introduces a normalization step by subtracting the last value in the sequence before applying the linear transformation. The aim of this transformation is to reduce bias from recent trends. Finally, we implemented the "DLinear" which decompose the time series into its trend and seasonal components. After the decomposition processed, linear layer is applied over each component. This decomposition allows the model to better capture and process different patterns within the time series. It's worth noting that these models are not inherently designed for multivariate to univariate prediction. To address this limitation, we added a final dense layer to aggregate the results.

\medskip

\textit{Recurrent Neural Networks (RNNs)} are often the default choice for time series forecasting and are a key component in more complex models like the Temporal Fusion Transformer. However, given recent criticisms of transformer models, we aimed to reassess the efficacy of RNNs for standard time series forecasting. While RNNs can capture complex non-linear patterns, they come with a computational cost due to their non-parallelizable sequential processing. 

\medskip

\textit{LSTM Encoder-Decoder} uses separate LSTM layers for encoding and decoding. The encoder (50 units) processes the input sequence and passes its final state to the decoder (50 units), which generates the output sequence. A final dense layer produces the predictions. 

\medskip

\textit{LSTM Shared Encoder-Decoder} Uses a single LSTM layer (units equal to the number of input features) shared between the encoding and decoding phases. We also make a comparison against two variations of GRU (Gated Recurrent Unit) networks 

\medskip

\textit{GRU Last Sequence}, consists of a two-layer GRU network with 10 units in each layer, where only the last output of the second layer is used for prediction. This approach focuses on capturing the final state of the sequence, which can represent the most relevant information for forecasting. The second variant, \textit{GRU Full Sequence}, also features a two-layer GRU network with 10 units per layer but differs by using the full output sequence of the second layer. The outputs are flattened and then used for prediction, allowing the model to leverage information from all time steps within the sequence, potentially capturing more intricate temporal patterns.

\medskip

\textit{Multilayer Perceptrons (MLPs)}: While not typically used for time series forecasting, we included MLPs as a fundamental deep learning baseline. As with linear regression, we flattened the inputs to accommodate the sequential data. We used linear activations to maintain comparability with our model. In previous tests, we observed that ReLU activations in small architectures like these often hindered performance in this scenario.\textit{Single-layer MLP}: Flattens the input and applies a single dense layer to produce the output.\textit{Two-layer MLP}: Flattens the input, applies a dense layer with (n\_ahead * 10 * 2) units, followed by another dense layer for the final prediction.

\medskip

\textit{TSMixer}: The TSMixer, which inspired our work, offers an alternative to RNNs and builds upon the ideas presented in "Are Transformers Effective for Time Series Forecasting?" It consists of the following key components: First, \textit{Reversible Instance Normalization} is employed to normalize the data while preserving the ability to revert to the original representation, which helps in maintaining stability during training. Next, a \textit{Temporal Linear layer} is utilized; it permutes the input dimensions and applies a dense layer, allowing for effective handling of temporal dependencies within the data. Finally, a \textit{Feature Linear layer} consisting of two dense layers with an activation function in between is integrated, enabling complex feature transformations and non-linear interactions. We tested four different variants;
We tested four variants: \textit{TSMixer-1} consists of a single block utilizing ReLU activation, while \textit{TSMixer-2} extends this to two blocks, also with ReLU activation, allowing for deeper feature extraction. In contrast, TSMixer-no-relu-1 features a single block with linear activation, and \textit{TSMixer-no-relu-2} uses two blocks with linear activation, offering an alternative approach without non-linear transformations. These configurations enable flexibility in handling different types of data and learning patterns. All TSMixer variants use layer normalization and a feed-forward dimension of 5. We included variants with linear activation to align more closely with TLN approach. As with the linear models mentioned earlier, we added a dense layer for result aggregation in multivariate to univariate prediction tasks.

\medskip

\textit{TLN (LinearNet)}:Our proposed Temporal Linear Network, implemented in two variants: \textit{LinearNet} which uses convolution operations and \textit{LinearNet-no-convolution} which does not embed convolution filter. Both variants use two hidden layers with default parameters for internal hidden size. This comprehensive set of models allows us to compare our proposed TLN against a wide range of approaches, from simple linear models to complex deep learning architectures, providing a thorough evaluation of its performance in various time series forecasting scenarios.

\subsection{Evaluation Metrics}

We used different evaluation metrics for our two tasks, chosen to facilitate comparison with existing literature and to provide the most meaningful and interpretable results for each specific dataset.

\medskip

\subsubsection{Task 1: Market Traded Notional Predictions}

For the first task, we used the R-squared (R²) score as our primary evaluation metric. The R² score is defined as:

\begin{equation}
R^2 = 1 - \frac{\sum_{i} (y_i - \hat{y}_i)^2}{\sum_{i} (y_i - \bar{y})^2},
\end{equation}
where $y_i$ are the true values, $\hat{y}_i$ are the predicted values, and $\bar{y}$ is the mean of the true values.

\medskip

\subsubsection{Task 2: ETTh1 Dataset}
For the second task, we used Mean Squared Error (MSE) as our primary evaluation metric. MSE is defined as:
\begin{equation}
MSE = \frac{1}{n} \sum_{i=1}^{n} (y_i - \hat{y}_i)^2,
\end{equation}
where $n$ is the number of predictions.

\medskip

\subsubsection{Rationale for Metric Selection}
The choice of different metrics for the two tasks serves multiple purposes:

\medskip

\textit{Consistency with existing literature}: By using R² for the financial task and MSE for the ETTh1 dataset, we maintain consistency with the metrics commonly used in these domains, facilitating direct comparisons with other papers.

\medskip

\textit{Readability of results}: In the BTC tasks, the R² scores typically fall in the range of 0.1 to 0.4. In this range of values, differences between models are more easily discernible with this metric. Conversely, for the ETTh1 tasks, R² scores would often exceed 0.9, compressing the visible differences between models. Using MSE for ETTh1 provides a clearer distinction between model performances.

\medskip

\textit{Scale-independent interpretation}: R² has the advantage of being scale-independent and directly interpretable as the proportion of variance explained by the model. This makes it particularly useful for the financial task where the absolute scale of errors may be less informative than the proportion of variance explained. It's important to note that R² and MSE are intrinsically related. Both metrics measure the same fundamental quantity - the discrepancy between predicted and actual values. The choice between them is primarily a matter of presentation and interpretation in the context of the specific task and data characteristics. The $\text{R}^2$ has an additional advantage in that it inherently accounts for the variance of the target variable. This makes it more immediately interpretable than MSE, as it provides a clear benchmark (with 1 being a perfect score and 0 indicating performance no better than always predicting the mean). By using these two complementary metrics across our tasks, we aim to provide a comprehensive and nuanced evaluation of model performance, balancing the needs for comparability with existing literature, readability of results, and meaningful interpretation in the context of each specific forecasting task.

\subsection{Results Analysis}
Our comprehensive experimental results, presented in the appendix for clarity, reveal key insights about model behaviors across different scenarios and demonstrate how TLN (referred to as LinearNet in the tables) provides an effective alternative to existing approaches. For Bitcoin volume prediction in univariate settings without time features, short-term predictions (1-step ahead) show similar performance across simpler models including linear regression, CLinear, NLinear, DLinear, MLP, TSMixer, and TLN. Surprisingly, RNNs perform poorly in these scenarios despite their prevalence in time-series forecasting applications. More significantly, deep learning linear models (CLinear, NLinear, and DLinear) and TSMixer show degraded performance with longer sequence lengths, while both linear regression and TLN maintain consistent performance regardless of sequence length.

\medskip

Regarding medium-term predictions (6-30 steps ahead), the performance hierarchy shifts substantially, with recurrent architectures demonstrating superior capabilities. However, our proposed TLN approach maintains stable performance while offering a significant advantage in computational efficiency. While RNN training times increase dramatically from 8 to 150 seconds as sequence length grows from 3 to 400, TLN training time remains nearly constant, increasing only marginally from 6.5 to 7.1 seconds. Long-term prediction scenarios particularly highlight TLN's strengths. While TSMixer occasionally matches its performance depending on the specific run, TLN consistently delivers stable results across all runs, contrasting sharply with the highly variable performance of other models. These results emphasize TLN's two key advantages: consistent handling of long sequences regardless of input characteristics and minimal training time compared to deep learning alternatives. Adding temporal features improves performance across all models, demonstrating the importance of this basic information. RNNs benefit significantly, showing stronger performance in both short and medium-term predictions without their previous short-term limitations. Even basic linear regression and ridge regression models perform well in these scenarios. TLN maintains consistent strong performance across configurations, with one notable exception: a simple 3-layer MLP achieves surprisingly good results for 360-step predictions with very short input sequences. However, TLN matches this performance when provided longer sequences, suggesting that while it may not extract optimal predictions from limited data, it effectively leverages longer sequences for reliable predictions.

\medskip

The transition to multivariate tasks, both with and without temporal features, reveals particularly interesting dynamics. As noted in previous research, increased input dimensionality often degrades model performance and complicates calibration, evident in our results tables. While some models become practically unusable (particularly TSMixer), TLN emerges as consistently superior across nearly all prediction horizons. This stability extends from its earlier demonstrated robustness to sequence length - as feature dimensionality increases, TLN maintains consistent performance while other models degrade. This advantage becomes clear when examining parameter counts: most models show dramatic parameter increases with added features, while TLN maintains a more modest growth rate, particularly compared to ridge regression or linear regression. Results on the ETTh1 dataset, which has previously highlighted the limitations of complex deep learning approaches, confirm earlier research showing simpler models outperforming RNNs. While individual models may excel in specific scenarios, TLN demonstrates unmatched consistency across all configurations. Frequently, other models achieve strong performance with certain sequence lengths but fail dramatically when input length changes. This consistency across varying conditions, combined with efficient training times and robust performance across input dimensions, establishes TLN as a practical advancement in time series prediction. This analysis underscores TLN's primary contributions: consistent performance across varying input conditions, efficient computational requirements, and reliable results across different prediction horizons. These characteristics make it particularly valuable for practical applications requiring robust performance across diverse scenarios.

\section{Comparing with Machine Learning Models}

As demonstrated in the learning tasks, linear models often prove not only sufficient but also more efficient for time series forecasting. These models should be considered as primary candidates before exploring non-linear models, which are frequently the default choice in deep learning literature. Having proposed a new architecture that essentially offers an innovative method for calibrating a linear model, we aim to highlight its distinctions from standard machine learning methods in terms of final model weights. To accomplish this, we reemployed the same two tasks from above, calibrating a model with a sequence length of 1000 for ETTh1 and 200 for the BTC task. The difference in sequence lengths is attributed to the higher number of input features in the BTC task, which significantly increases the complexity of standard model calibration.For both tasks, we trained the model to predict 128 steps ahead, utilizing multivariate inputs that include time information. We then leveraged the TLN's capacity to be rewritten as a linear equation and visualized the weights in a heatmap. The heatmap axes represent the features and the sequence length. We compare the weights using TLN, linear regression, ridge \cite{hoerl1970ridge}, lasso \cite{tibshirani1996regression}, and ElasticNet \cite{zou2005regularization} .
\begin{figure}[H]
    \centering
    \includegraphics[width=0.4\textwidth]{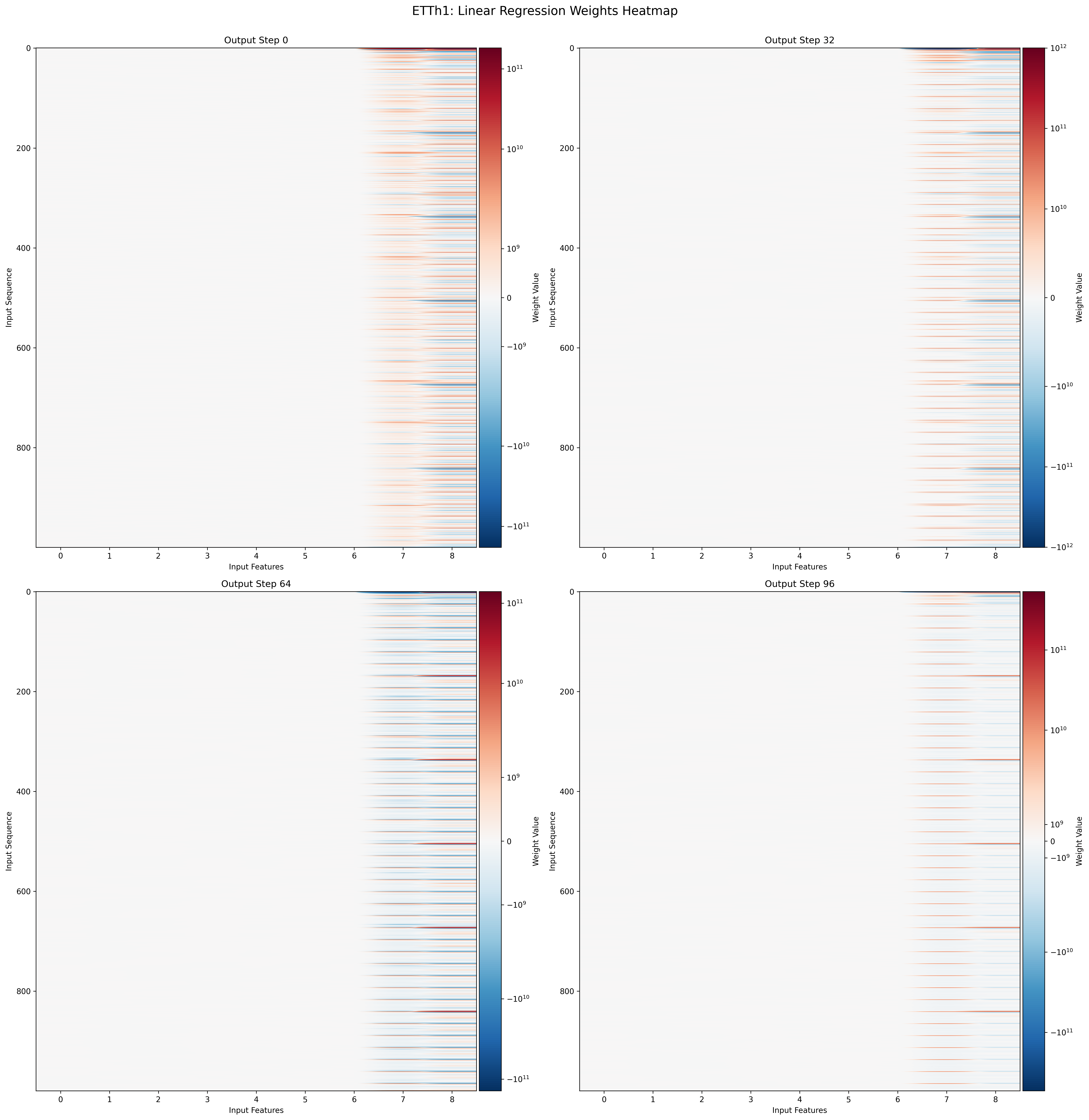}
    \caption{Linear Regression weights for ETTh1 task}
    \label{fig:etth1_linreg_weights}
\end{figure}
In the Linear Regression model, the weights appear to be disproportionately concentrated on the last two features, which represent temporal information. Notably, the magnitude of these values is exceptionally high, considering our targets are scaled to the [0,1] range.
\begin{figure}[H]
    \centering
    \includegraphics[width=0.4\textwidth]{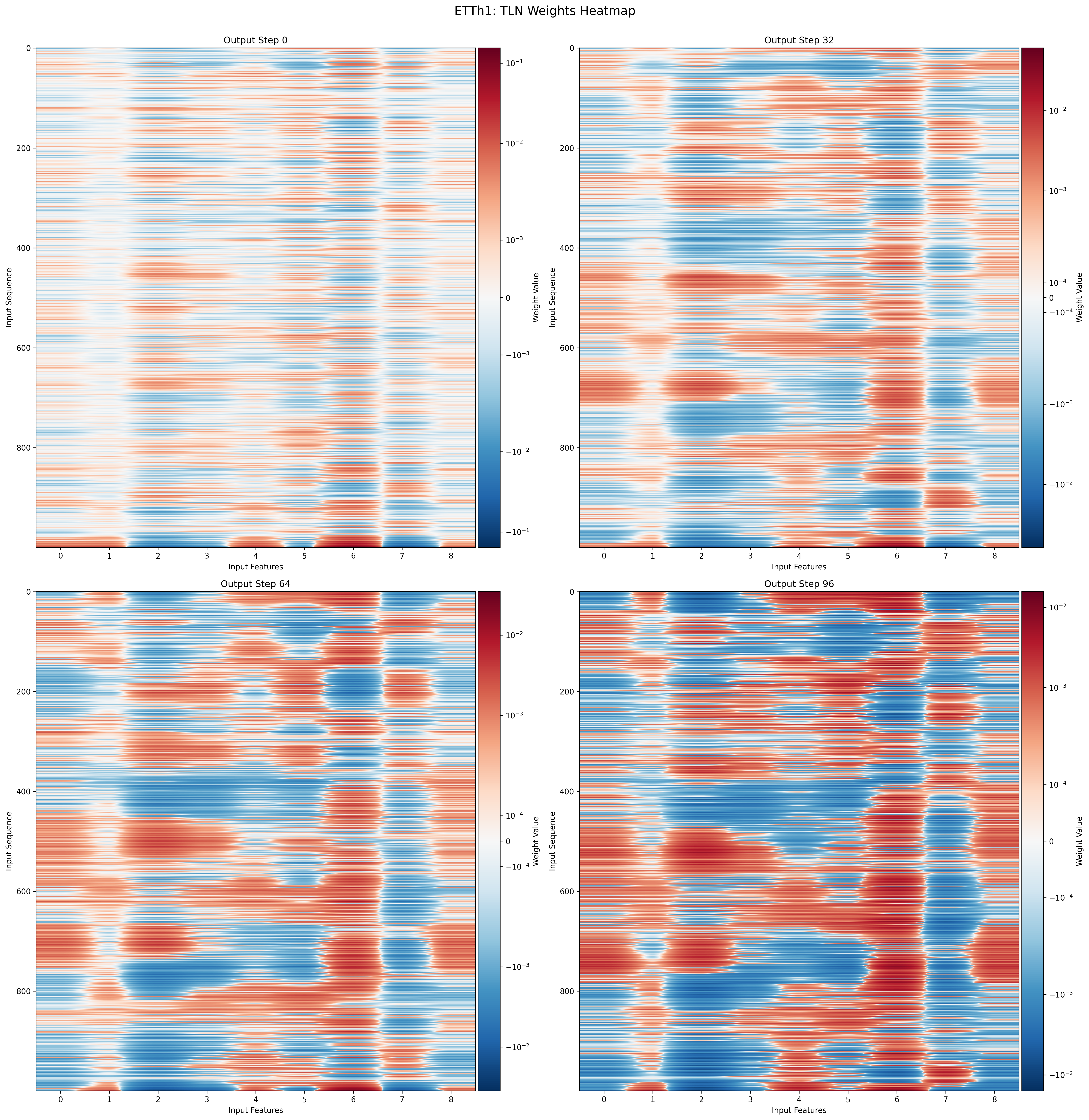}
    \caption{TLN weights for ETTh1 task}
    \label{fig:etth1_tln_weights}
\end{figure}
In contrast, the TLN demonstrates a more balanced distribution of weights across all features, albeit with varying magnitudes. An interesting and expected observation is that weights for 1-step-ahead predictions are heavily biased towards recent events, while longer-term predictions incorporate more information from earlier elements in the sequences.
\begin{figure}[H]
    \centering
    \includegraphics[width=0.4\textwidth]{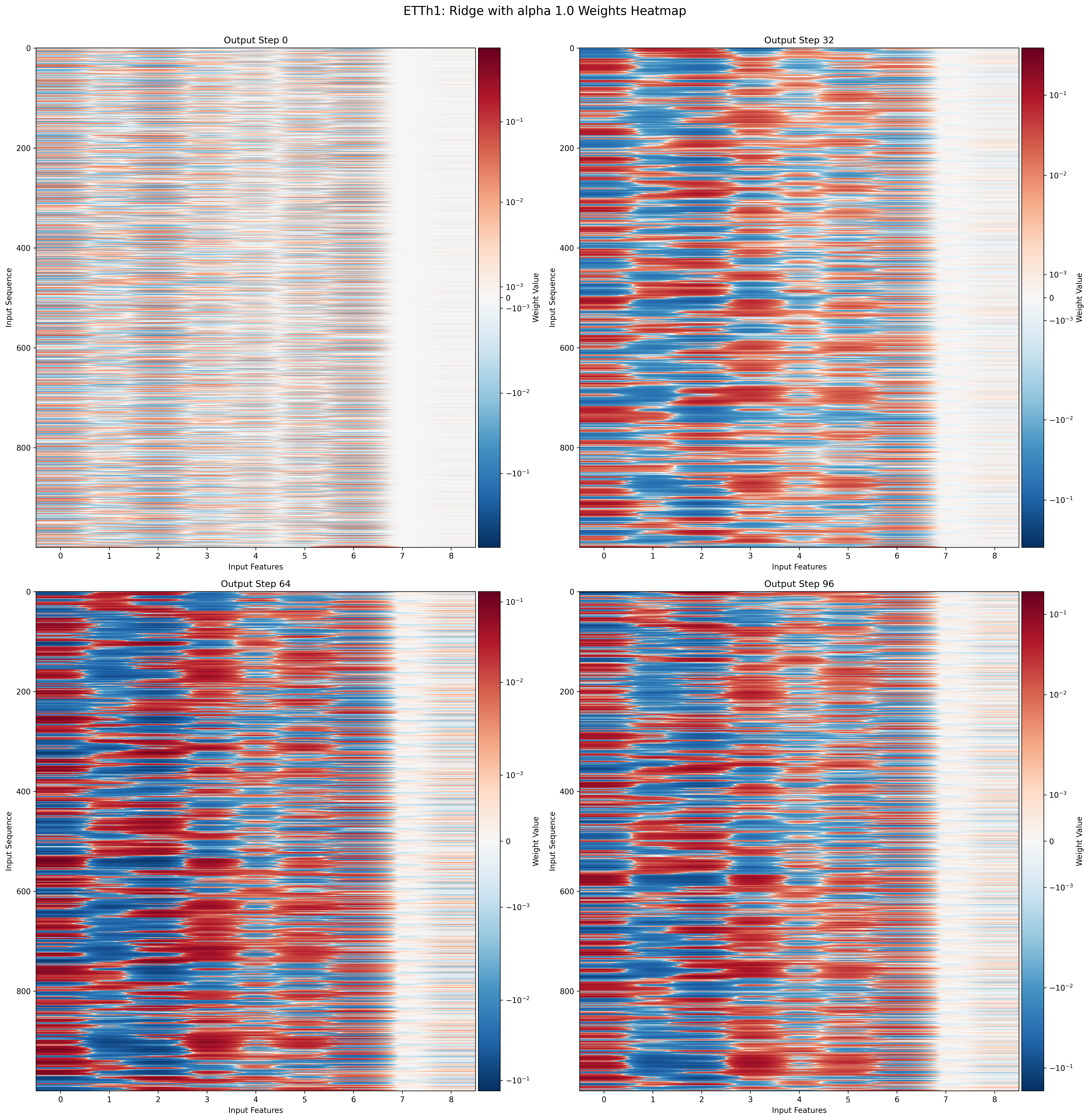}
    \caption{Ridge Regression weights for ETTh1 task (alpha = 1.0)}
    \label{fig:etth1_ridge_weights}
\end{figure}
The Ridge Regression model shows less dominance of temporal information compared to Linear Regression. However, it lacks the TLN's characteristic of emphasizing recent values for short-term predictions, which is a logical consequence of the applied weight regularization.
\begin{figure}[H]
    \centering
    \includegraphics[width=0.4\textwidth]{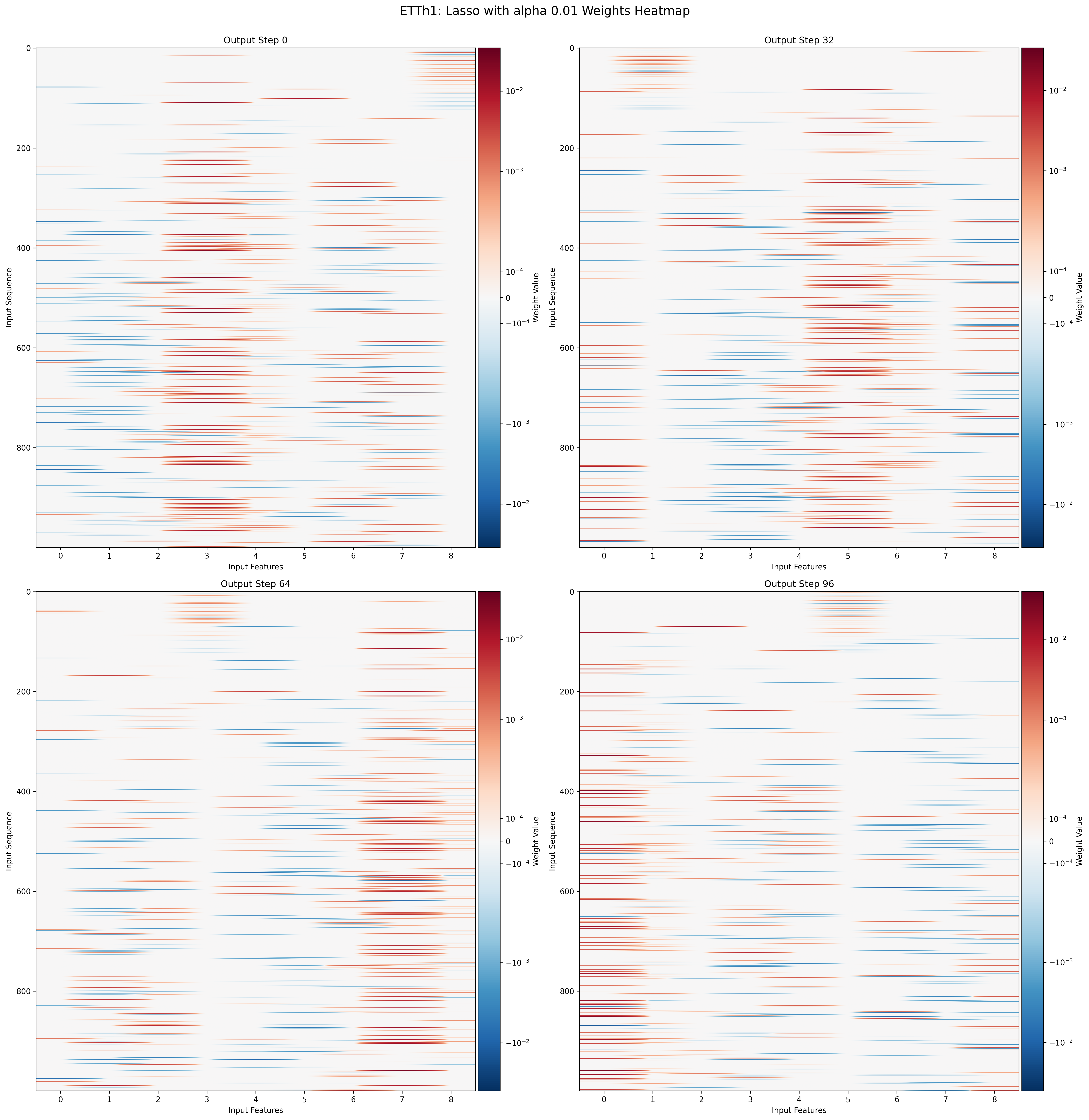}
    \caption{Lasso Regression weights for ETTh1 task (alpha = 0.01)}
    \label{fig:etth1_lasso_weights}
\end{figure}

\begin{figure}[H]
    \centering
    \includegraphics[width=0.4\textwidth]{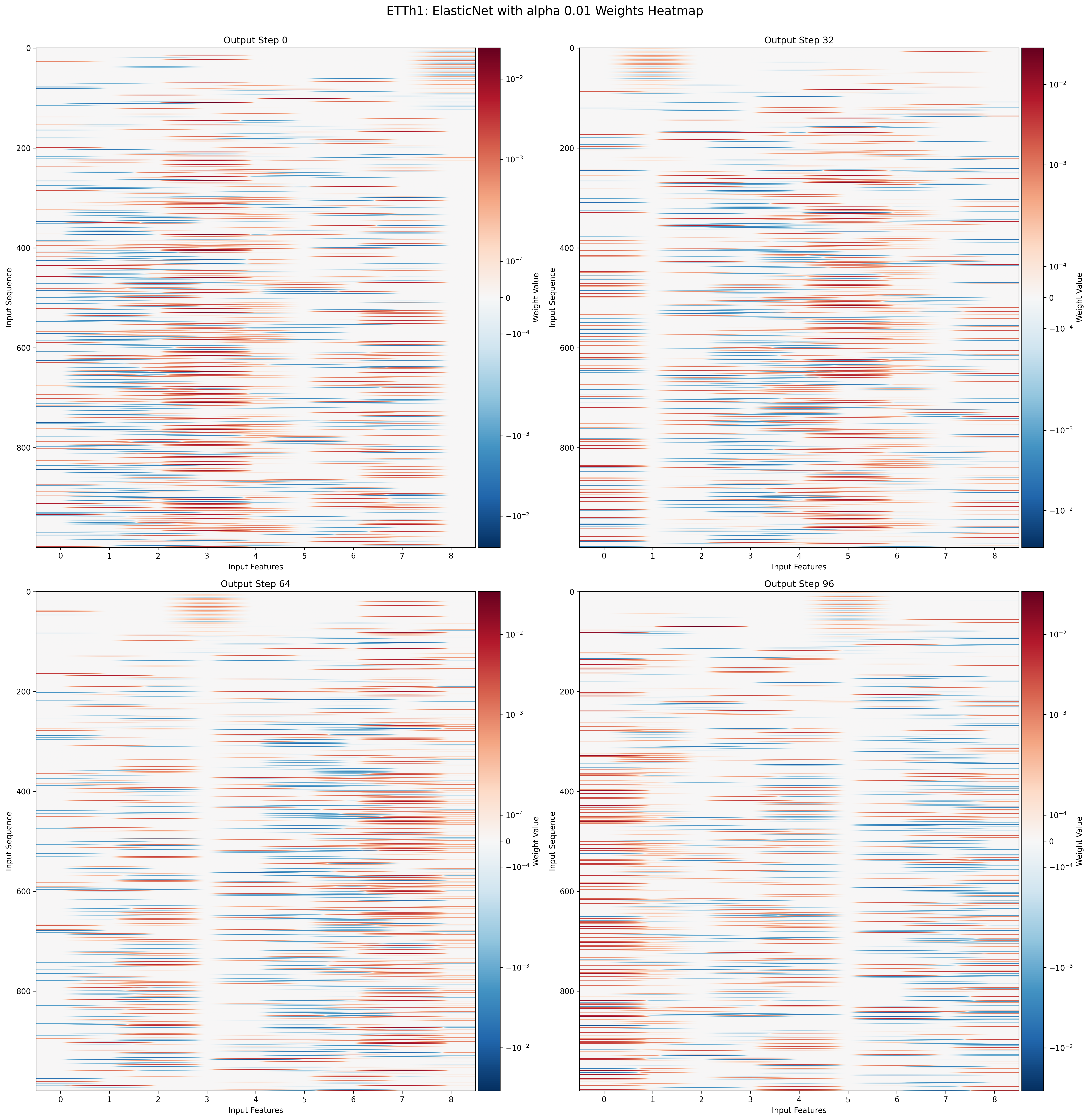}
    \caption{ElasticNet Regression weights for ETTh1 task (alpha = 0.01)}
    \label{fig:etth1_elastic_weights}
\end{figure}
Lasso and ElasticNet models exhibit distinct characteristics. While both utilize a broader range of parameters than Linear Regression, they demonstrate their strength in feature selection by setting many weights to zero. Interestingly, temporal information is only used at specific points in the sequence, which is logical given the continuous nature of the series. In practice, only one time feature would be necessary, although we included multiple for matrix formatting purposes.

\medskip

For the ETTh1 tasks, Lasso and ElasticNet outperform the TLN, while Linear Regression and Ridge fall short. This aligns with our observations across all models: Linear Regression excelled in ETTh1 tasks but became less efficient with increased inputs, while Lasso and ElasticNet's corrective effects enhanced performance. Conversely, for the BTC tasks, the TLN outperformed other models, underscoring its value as a linear alternative that operates distinctly from traditional approaches. A noteworthy observation from our tests is the TLN's robustness in handling larger inputs. While standard ML models can become computationally intensive to calibrate with low alpha values or numerous features, the TLN consistently performs well. Although it may require more time for smaller datasets, it demonstrates superior scalability when calibrating on substantially larger inputs.
\section{Conclusion}
In this paper, we introduced the Temporal Linear Network (TLN), a novel architecture that bridges the gap between simple linear models and complex deep learning approaches for time series forecasting. Our extensive experimental results across multiple tasks and configurations demonstrate several key findings.

First, TLN achieves robust performance across varying input conditions while maintaining the interpretability advantages of linear models. The architecture's ability to be transformed into an equivalent linear model provides transparency through explicit weight relationships. The successive transformations applied in the model (alternating between feature and temporal dimensions) create structured weight dependencies that act as an implicit regularization mechanism. This internal structure results in a more compact representation compared to standard linear models, requiring fewer parameters while maintaining or improving performance.

\medskip

The reduced parameter space, combined with the structured transformations, explains both the model's stability across different sequence lengths and its computational efficiency. Where a standard linear model would require a full \(S \times F \times S' \times F'\) weight matrix for a transformation from an input of shape $(S,F)$ to an output of shape $(S',F')$, our approach decomposes this into a series of smaller transformations. This decomposition not only reduces the memory footprint but also results in more efficient matrix multiplications during both training and inference.

\medskip

Our comparative analysis with traditional machine learning methods reveals that TLN's weight distributions show more balanced patterns than standard linear regression approaches. The model's performance on both the ETTh1 dataset and Bitcoin volume prediction tasks demonstrates consistent accuracy across different prediction horizons, from short-term to long-term forecasting. This consistency is particularly noteworthy in multivariate settings, where many competing approaches show significant degradation in performance.

\medskip

The key advantage of TLN lies in its ability to maintain the interpretability of linear models while achieving superior performance through its structured parameter space. This is accomplished without the computational overhead typically associated with deep learning approaches, making it particularly valuable for real-world applications where both performance and interpretability are crucial.

\medskip

Our results suggest that TLN represents a significant advancement in time series forecasting, offering an optimal balance between model complexity and interpretability. Its consistent performance across various scenarios, combined with its computational efficiency and theoretical guarantees of linearity, makes it a valuable addition to the existing toolkit of time series forecasting methods.

\bibliographystyle{IEEEtran}
\bibliography{bib}

\appendices
\onecolumn

\section{BTC notionnals prediction tasks results}
\subsection{R2 scores}

\begin{table}[H][H]
\caption{Univariate without Time: R2 Score (x1e-2)}
\label{tab:univariate_no_time_r2_btc}
\scriptsize  
\setlength{\tabcolsep}{2pt}  
\begin{tabular}{llllllllllllllllllll}
\toprule
 &  & \shortstack{Linear \\ Regression} & Ridge & \shortstack{LSTM \\ Encoder \\ Decoder} & \shortstack{LSTM \\ Shared \\ Encoder \\ Decoder} & \shortstack{GRU \\ Last \\ Sequence} & \shortstack{GRU \\ Full \\ Sequence} & \shortstack{MLP \\ 1 \\ Layer \\ linear} & \shortstack{MLP \\ 2 \\ Layer \\ linear} & \shortstack{MLP \\ 3 \\ Layer \\ ReLU} & LinearNet & \shortstack{LinearNet \\ no \\ conv} & CLinear & NLinear & DLinear & \shortstack{TSMixer \\ 1} & \shortstack{TSMixer \\ 2} & \shortstack{TSMixer \\ no \\ relu \\ 1} & \shortstack{TSMixer \\ no \\ relu \\ 2} \\
\shortstack{n \\ ahead} & \shortstack{sequence \\ length} &  &  &  &  &  &  &  &  &  &  &  &  &  &  &  &  &  &  \\
\midrule
\multirow[t]{7}{*}{1} & 3 & 41.15 & 41.02 & \textcolor{blue}{39.68} & \textcolor{blue}{37.71} & 35.88 & 40.24 & 40.32 & 40.74 & \textcolor{blue}{39.92} & 40.92 & 40.69 & 40.52 & 40.89 & 40.72 & \textcolor{blue}{41.26} & \textcolor{blue}{\textbf{\underline{42.22}}} & \textcolor{blue}{40.84} & 32.43 \\
 & 10 & \underline{41.08} & 40.94 & 37.56 & 32.44 & \textcolor{blue}{40.46} & 39.55 & 40.57 & 40.73 & 33.75 & 40.32 & 40.70 & 40.65 & 41.02 & 40.79 & 32.19 & 39.95 & 33.08 & 39.69 \\
 & 20 & \underline{41.51} & 41.37 & 33.50 & 31.87 & 37.38 & 38.41 & 41.13 & \textcolor{blue}{41.37} & 33.90 & 41.17 & 41.04 & 40.98 & \textcolor{blue}{41.19} & 40.97 & 40.27 & 29.82 & 40.67 & \textcolor{blue}{40.97} \\
 & 50 & \textcolor{blue}{\underline{41.64}} & \textcolor{blue}{41.51} & 37.01 & 28.10 & 38.37 & 40.15 & \textcolor{blue}{41.41} & 41.01 & 32.73 & \textcolor{blue}{41.52} & \textcolor{blue}{41.36} & 40.99 & 26.60 & 40.75 & 39.16 & 35.00 & 30.87 & 29.07 \\
 & 100 & \underline{41.57} & 41.44 & 31.96 & 33.51 & 38.69 & \textcolor{blue}{40.50} & 41.05 & 40.88 & 30.02 & 41.36 & 40.90 & 40.79 & 31.78 & \textcolor{blue}{40.99} & 33.15 & 20.71 & 21.85 & 30.14 \\
\bottomrule
\end{tabular}
\end{table}

\begin{table}[H][H]
\caption{Univariate with Time: R2 Score (x1e-2)}
\label{tab:univariate_with_time_r2_btc}
\scriptsize  
\setlength{\tabcolsep}{2pt}  
\begin{tabular}{llllllllllllllllllll}
\toprule
 &  & \shortstack{Linear \\ Regression} & Ridge & \shortstack{LSTM \\ Encoder \\ Decoder} & \shortstack{LSTM \\ Shared \\ Encoder \\ Decoder} & \shortstack{GRU \\ Last \\ Sequence} & \shortstack{GRU \\ Full \\ Sequence} & \shortstack{MLP \\ 1 \\ Layer \\ linear} & \shortstack{MLP \\ 2 \\ Layer \\ linear} & \shortstack{MLP \\ 3 \\ Layer \\ ReLU} & LinearNet & \shortstack{LinearNet \\ no \\ conv} & CLinear & NLinear & DLinear & \shortstack{TSMixer \\ 1} & \shortstack{TSMixer \\ 2} & \shortstack{TSMixer \\ no \\ relu \\ 1} & \shortstack{TSMixer \\ no \\ relu \\ 2} \\
\shortstack{n \\ ahead} & \shortstack{sequence \\ length} &  &  &  &  &  &  &  &  &  &  &  &  &  &  &  &  &  &  \\
\midrule
\multirow[t]{7}{*}{1} & 3 & 41.37 & 41.24 & 39.07 & 32.98 & 40.38 & 38.88 & 40.80 & 41.12 & 40.05 & 41.17 & 40.57 & 3.31 & 41.04 & 41.18 & 42.00 & \textcolor{blue}{\underline{42.62}} & 34.95 & 35.02 \\
 & 10 & 41.63 & 41.47 & \textcolor{blue}{\underline{41.67}} & 34.71 & 38.31 & 39.75 & 41.00 & 41.23 & 38.53 & 41.34 & 40.47 & 0.82 & 41.25 & 8.04 & 39.64 & 40.05 & \textcolor{blue}{40.52} & \textcolor{blue}{41.62} \\
 & 20 & \underline{42.80} & 42.66 & 37.33 & 35.81 & 40.38 & 40.78 & 41.61 & 41.84 & \textcolor{blue}{40.73} & \textcolor{blue}{42.09} & \textcolor{blue}{41.00} & \textcolor{blue}{41.44} & \textcolor{blue}{41.48} & \textcolor{blue}{41.54} & \textcolor{blue}{42.72} & 41.35 & 40.33 & 30.41 \\
 & 50 & \textcolor{blue}{\textbf{\underline{42.84}}} & \textcolor{blue}{42.70} & 36.42 & 34.63 & \textcolor{blue}{42.07} & 40.47 & \textcolor{blue}{41.81} & \textcolor{blue}{42.10} & 39.44 & 41.88 & 40.79 & 28.73 & 32.99 & 41.25 & 27.92 & 29.52 & 28.17 & 37.15 \\
 & 100 & \underline{42.64} & 42.51 & 40.83 & \textcolor{blue}{37.46} & 37.90 & 40.48 & 40.29 & 40.65 & 38.69 & 41.80 & 38.76 & 40.86 & 33.08 & 8.81 & 25.40 & 30.25 & 39.55 & 27.18 \\
\bottomrule
\end{tabular}
\end{table}

\begin{table}[H][H]
\caption{Multivariate without Time: R2 Score (x1e-2)}
\label{tab:multivariate_no_time_r2_btc}
\scriptsize  
\setlength{\tabcolsep}{2pt}  
\begin{tabular}{llllllllllllllllllll}
\toprule
 &  & \shortstack{Linear \\ Regression} & Ridge & \shortstack{LSTM \\ Encoder \\ Decoder} & \shortstack{LSTM \\ Shared \\ Encoder \\ Decoder} & \shortstack{GRU \\ Last \\ Sequence} & \shortstack{GRU \\ Full \\ Sequence} & \shortstack{MLP \\ 1 \\ Layer \\ linear} & \shortstack{MLP \\ 2 \\ Layer \\ linear} & \shortstack{MLP \\ 3 \\ Layer \\ ReLU} & LinearNet & \shortstack{LinearNet \\ no \\ conv} & CLinear & NLinear & DLinear & \shortstack{TSMixer \\ 1} & \shortstack{TSMixer \\ 2} & \shortstack{TSMixer \\ no \\ relu \\ 1} & \shortstack{TSMixer \\ no \\ relu \\ 2} \\
\shortstack{n \\ ahead} & \shortstack{sequence \\ length} &  &  &  &  &  &  &  &  &  &  &  &  &  &  &  &  &  &  \\
\midrule
\multirow[t]{7}{*}{1} & 3 & \textcolor{blue}{40.80} & \textcolor{blue}{40.51} & \textcolor{blue}{41.14} & 39.82 & 38.69 & \textcolor{blue}{40.66} & \textcolor{blue}{38.80} & \textcolor{blue}{40.59} & \textcolor{blue}{34.61} & 39.37 & 39.13 & 40.06 & 40.15 & 39.91 & \textcolor{blue}{41.18} & \textcolor{blue}{\textbf{\underline{42.47}}} & 32.76 & 33.81 \\
 & 10 & 40.16 & 39.90 & 39.25 & 37.56 & 39.31 & 38.30 & 37.52 & 39.57 & 30.22 & 40.27 & 40.03 & 40.46 & \textcolor{blue}{\underline{41.25}} & 39.86 & 39.59 & 35.81 & \textcolor{blue}{37.25} & 38.74 \\
 & 20 & 39.80 & 39.69 & 39.98 & 39.33 & 39.10 & 38.08 & 37.21 & 38.36 & 33.45 & \textcolor{blue}{\underline{41.22}} & 40.58 & 40.51 & 38.06 & 41.01 & 19.10 & 34.98 & 36.36 & \textcolor{blue}{38.90} \\
 & 50 & 38.09 & 38.57 & 39.49 & \textcolor{blue}{40.29} & 40.22 & 38.00 & 36.43 & 36.66 & 32.51 & 41.08 & \textcolor{blue}{40.90} & \textcolor{blue}{40.78} & 40.59 & \textcolor{blue}{\underline{41.15}} & 31.47 & 31.29 & 32.08 & 33.91 \\
 & 100 & 35.61 & 37.06 & 39.54 & 39.91 & 40.46 & 39.54 & 33.15 & 34.42 & 28.88 & \underline{41.12} & 40.86 & 40.49 & 40.42 & 40.97 & 31.11 & 26.25 & 16.79 & 33.36 \\
\bottomrule
\end{tabular}

\end{table}

\begin{table}[H][H]
\caption{Multivariate with Time: R2 Score (x1e-2)}
\label{tab:multivariate_with_time_r2_btc}
\scriptsize  
\setlength{\tabcolsep}{2pt}  
\begin{tabular}{llllllllllllllllllll}
\toprule
 &  & \shortstack{Linear \\ Regression} & Ridge & \shortstack{LSTM \\ Encoder \\ Decoder} & \shortstack{LSTM \\ Shared \\ Encoder \\ Decoder} & \shortstack{GRU \\ Last \\ Sequence} & \shortstack{GRU \\ Full \\ Sequence} & \shortstack{MLP \\ 1 \\ Layer \\ linear} & \shortstack{MLP \\ 2 \\ Layer \\ linear} & \shortstack{MLP \\ 3 \\ Layer \\ ReLU} & LinearNet & \shortstack{LinearNet \\ no \\ conv} & CLinear & NLinear & DLinear & \shortstack{TSMixer \\ 1} & \shortstack{TSMixer \\ 2} & \shortstack{TSMixer \\ no \\ relu \\ 1} & \shortstack{TSMixer \\ no \\ relu \\ 2} \\
\shortstack{n \\ ahead} & \shortstack{sequence \\ length} &  &  &  &  &  &  &  &  &  &  &  &  &  &  &  &  &  &  \\
\midrule
\multirow[t]{7}{*}{1} & 3 & 40.83 & 40.52 & \textcolor{blue}{\underline{40.95}} & \textcolor{blue}{39.93} & 37.19 & 37.11 & 38.90 & \textcolor{blue}{40.30} & \textcolor{blue}{37.36} & 39.17 & 39.13 & 39.55 & \textcolor{blue}{39.80} & 39.58 & \textcolor{blue}{37.00} & \textcolor{blue}{35.86} & 38.48 & 40.64 \\
 & 10 & 40.42 & 40.13 & 39.45 & 37.35 & \textcolor{blue}{\underline{41.62}} & 36.15 & 38.23 & 40.02 & 30.64 & 39.53 & 40.45 & 40.32 & 38.97 & \textcolor{blue}{40.55} & 36.08 & 34.94 & \textcolor{blue}{39.75} & 38.05 \\
 & 20 & \textcolor{blue}{41.13} & \textcolor{blue}{40.96} & 40.25 & 37.29 & 40.07 & 38.04 & \textcolor{blue}{39.00} & 40.27 & 33.35 & \underline{41.34} & \textcolor{blue}{41.20} & \textcolor{blue}{41.21} & 38.93 & 38.49 & 22.56 & 23.91 & 36.31 & \textcolor{blue}{41.01} \\
 & 50 & 39.33 & 39.76 & 39.87 & 39.22 & 38.96 & \textcolor{blue}{38.53} & 35.99 & 38.28 & 31.67 & \textcolor{blue}{\textbf{\underline{42.09}}} & 40.70 & 41.07 & 29.37 & 32.99 & 35.43 & 21.35 & 35.71 & 21.31 \\
 & 100 & 36.38 & 37.74 & 39.91 & 38.64 & 40.16 & 37.30 & 33.86 & 34.44 & 28.05 & 40.57 & 39.51 & \underline{40.88} & 38.15 & 35.04 & 22.35 & 26.50 & 34.10 & 35.83 \\
\bottomrule
\end{tabular}

\end{table}

\subsection{Training Times}

\begin{table}[H][H]
\caption{Univariate without Time: Training time}
\label{tab:univariate_no_time_train_time_btc}
\scriptsize  
\setlength{\tabcolsep}{2pt}  
\begin{tabular}{llllllllllllllllllll}
\toprule
 &  & \shortstack{Linear \\ Regression} & Ridge & \shortstack{LSTM \\ Encoder \\ Decoder} & \shortstack{LSTM \\ Shared \\ Encoder \\ Decoder} & \shortstack{GRU \\ Last \\ Sequence} & \shortstack{GRU \\ Full \\ Sequence} & \shortstack{MLP \\ 1 \\ Layer \\ linear} & \shortstack{MLP \\ 2 \\ Layer \\ linear} & \shortstack{MLP \\ 3 \\ Layer \\ ReLU} & LinearNet & \shortstack{LinearNet \\ no \\ conv} & CLinear & NLinear & DLinear & \shortstack{TSMixer \\ 1} & \shortstack{TSMixer \\ 2} & \shortstack{TSMixer \\ no \\ relu \\ 1} & \shortstack{TSMixer \\ no \\ relu \\ 2} \\
\shortstack{n \\ ahead} & \shortstack{sequence \\ length} &  &  &  &  &  &  &  &  &  &  &  &  &  &  &  &  &  &  \\
\midrule
\multirow[t]{7}{*}{1} & 3 & \textcolor{blue}{0.0} & \textcolor{blue}{\textbf{\underline{0.0}}} & \textcolor{blue}{6.6} & \textcolor{blue}{9.8} & \textcolor{blue}{7.8} & \textcolor{blue}{7.1} & 5.4 & 3.3 & \textcolor{blue}{4.7} & 7.2 & \textcolor{blue}{6.1} & \textcolor{blue}{3.7} & \textcolor{blue}{4.7} & 4.2 & 10.1 & 10.9 & 11.3 & 15.1 \\
 & 10 & 0.0 & \underline{0.0} & 7.6 & 13.6 & 8.8 & 9.2 & 4.4 & \textcolor{blue}{3.2} & 5.3 & 7.6 & 6.4 & 4.4 & 4.9 & 4.8 & 13.4 & 10.8 & 15.0 & 9.1 \\
 & 20 & 0.0 & \underline{0.0} & 9.0 & 12.5 & 12.8 & 16.6 & 4.8 & 4.1 & 4.9 & 7.4 & 7.3 & 5.0 & 5.4 & 4.4 & 10.4 & 14.6 & \textcolor{blue}{9.7} & \textcolor{blue}{8.9} \\
 & 50 & 0.0 & \underline{0.0} & 13.1 & 25.0 & 25.4 & 26.3 & 4.3 & 3.4 & 5.7 & \textcolor{blue}{7.0} & 7.3 & 4.9 & 6.9 & \textcolor{blue}{4.0} & 8.9 & 11.3 & 10.8 & 10.6 \\
 & 100 & 0.1 & \underline{0.0} & 23.7 & 60.1 & 38.4 & 30.0 & 4.4 & 3.8 & 5.7 & 8.2 & 8.4 & 7.0 & 8.8 & 4.1 & 9.1 & 11.9 & 12.1 & 12.1 \\
 & 200 & nan & nan & 32.5 & 81.0 & 50.7 & 72.9 & \textcolor{blue}{4.0} & \underline{3.6} & 5.9 & 8.0 & 7.6 & 5.3 & 5.1 & 5.6 & 8.1 & 10.6 & 12.0 & 11.3 \\
\bottomrule
\end{tabular}

\end{table}

\begin{table}[H][H]
\caption{Univariate with Time: Training time}
\label{tab:univariate_with_time_train_time_btc}
\scriptsize  
\setlength{\tabcolsep}{2pt}  
\begin{tabular}{llllllllllllllllllll}
\toprule
 &  & \shortstack{Linear \\ Regression} & Ridge & \shortstack{LSTM \\ Encoder \\ Decoder} & \shortstack{LSTM \\ Shared \\ Encoder \\ Decoder} & \shortstack{GRU \\ Last \\ Sequence} & \shortstack{GRU \\ Full \\ Sequence} & \shortstack{MLP \\ 1 \\ Layer \\ linear} & \shortstack{MLP \\ 2 \\ Layer \\ linear} & \shortstack{MLP \\ 3 \\ Layer \\ ReLU} & LinearNet & \shortstack{LinearNet \\ no \\ conv} & CLinear & NLinear & DLinear & \shortstack{TSMixer \\ 1} & \shortstack{TSMixer \\ 2} & \shortstack{TSMixer \\ no \\ relu \\ 1} & \shortstack{TSMixer \\ no \\ relu \\ 2} \\
\shortstack{n \\ ahead} & \shortstack{sequence \\ length} &  &  &  &  &  &  &  &  &  &  &  &  &  &  &  &  &  &  \\
\midrule
\multirow[t]{7}{*}{1} & 3 & \textcolor{blue}{0.0} & \textcolor{blue}{\textbf{\underline{0.0}}} & \textcolor{blue}{7.5} & \textcolor{blue}{9.3} & \textcolor{blue}{7.5} & \textcolor{blue}{8.5} & 8.6 & 4.7 & \textcolor{blue}{4.9} & \textcolor{blue}{7.8} & 7.0 & 7.6 & 8.1 & 11.8 & 15.7 & 12.6 & 17.4 & 13.6 \\
 & 10 & 0.0 & \underline{0.0} & 7.5 & 16.5 & 12.1 & 9.9 & 5.6 & \textcolor{blue}{4.3} & 6.0 & 8.9 & 7.2 & \textcolor{blue}{4.3} & 8.4 & 6.0 & 15.9 & \textcolor{blue}{11.9} & 17.3 & 17.4 \\
 & 20 & 0.0 & \underline{0.0} & 11.1 & 12.4 & 16.0 & 14.9 & 8.5 & 5.5 & 6.0 & 8.6 & \textcolor{blue}{6.1} & 8.7 & 7.0 & 12.6 & 16.3 & 18.6 & 21.1 & 15.8 \\
 & 50 & 0.1 & \underline{0.1} & 15.4 & 23.6 & 22.1 & 35.6 & 7.4 & 6.6 & 6.3 & 9.2 & 10.3 & 9.4 & 9.2 & 8.5 & 19.3 & 16.2 & 19.1 & 12.2 \\
 & 100 & 0.6 & \underline{0.2} & 24.1 & 51.8 & 62.4 & 53.6 & 7.7 & 5.5 & 7.4 & 11.2 & 9.9 & 9.3 & 6.9 & 5.0 & 22.7 & 21.5 & \textcolor{blue}{16.1} & 15.0 \\
 & 200 & nan & nan & 41.0 & 74.9 & 81.2 & 113.3 & 8.2 & 5.6 & 7.2 & 11.7 & 9.7 & \underline{5.0} & 8.4 & 8.2 & \textcolor{blue}{13.9} & 14.5 & 18.4 & 12.1 \\
\bottomrule
\end{tabular}

\end{table}

\begin{table}[H][H]
\caption{Multivariate without Time: Training time}
\label{tab:multivariate_no_time_train_time_btc}
\scriptsize  
\setlength{\tabcolsep}{2pt}  
\begin{tabular}{llllllllllllllllllll}
\toprule
 &  & \shortstack{Linear \\ Regression} & Ridge & \shortstack{LSTM \\ Encoder \\ Decoder} & \shortstack{LSTM \\ Shared \\ Encoder \\ Decoder} & \shortstack{GRU \\ Last \\ Sequence} & \shortstack{GRU \\ Full \\ Sequence} & \shortstack{MLP \\ 1 \\ Layer \\ linear} & \shortstack{MLP \\ 2 \\ Layer \\ linear} & \shortstack{MLP \\ 3 \\ Layer \\ ReLU} & LinearNet & \shortstack{LinearNet \\ no \\ conv} & CLinear & NLinear & DLinear & \shortstack{TSMixer \\ 1} & \shortstack{TSMixer \\ 2} & \shortstack{TSMixer \\ no \\ relu \\ 1} & \shortstack{TSMixer \\ no \\ relu \\ 2} \\
\shortstack{n \\ ahead} & \shortstack{sequence \\ length} &  &  &  &  &  &  &  &  &  &  &  &  &  &  &  &  &  &  \\
\midrule
\multirow[t]{7}{*}{1} & 3 & \textcolor{blue}{0.0} & \textcolor{blue}{\textbf{\underline{0.0}}} & \textcolor{blue}{6.9} & \textcolor{blue}{7.2} & \textcolor{blue}{7.4} & \textcolor{blue}{7.6} & \textcolor{blue}{3.9} & 3.9 & 4.6 & \textcolor{blue}{7.9} & \textcolor{blue}{6.7} & 6.3 & \textcolor{blue}{4.7} & 6.0 & \textcolor{blue}{12.9} & 13.2 & 15.0 & 12.7 \\
 & 10 & 0.2 & \underline{0.2} & 7.0 & 7.6 & 10.5 & 9.2 & 4.0 & 4.3 & 4.4 & 8.4 & 7.7 & 6.0 & 5.2 & \textcolor{blue}{5.9} & 16.4 & 13.2 & 15.2 & \textcolor{blue}{10.6} \\
 & 20 & 0.9 & \underline{0.2} & 10.0 & 9.7 & 13.6 & 12.9 & 4.2 & \textcolor{blue}{3.8} & 4.8 & 9.7 & 7.6 & 5.4 & 5.8 & 6.4 & 17.6 & \textcolor{blue}{11.9} & 15.1 & 12.4 \\
 & 50 & 4.3 & \underline{0.6} & 14.8 & 14.7 & 19.0 & 29.5 & 5.0 & 4.3 & 5.2 & 8.3 & 8.3 & 8.0 & 6.3 & 7.6 & 18.1 & 12.3 & 17.7 & 12.5 \\
 & 100 & 11.0 & \underline{1.0} & 18.0 & 22.3 & 41.9 & 43.6 & 5.1 & 4.7 & \textcolor{blue}{4.4} & 9.9 & 8.8 & \textcolor{blue}{5.4} & 7.3 & 6.4 & 19.1 & 12.8 & 16.2 & 12.8 \\
 & 200 & nan & nan & 37.3 & 40.1 & 76.3 & 73.4 & 5.0 & 6.1 & \underline{4.9} & 11.5 & 8.9 & 6.3 & 6.7 & 8.3 & 15.4 & 12.1 & \textcolor{blue}{14.6} & 12.1 \\
\bottomrule
\end{tabular}
\end{table}

\begin{table}[H][H]
\caption{Multivariate with Time: Training time}
\label{tab:multivariate_with_time_train_time_btc}
\scriptsize  
\setlength{\tabcolsep}{2pt}  
\begin{tabular}{llllllllllllllllllll}
\toprule
 &  & \shortstack{Linear \\ Regression} & Ridge & \shortstack{LSTM \\ Encoder \\ Decoder} & \shortstack{LSTM \\ Shared \\ Encoder \\ Decoder} & \shortstack{GRU \\ Last \\ Sequence} & \shortstack{GRU \\ Full \\ Sequence} & \shortstack{MLP \\ 1 \\ Layer \\ linear} & \shortstack{MLP \\ 2 \\ Layer \\ linear} & \shortstack{MLP \\ 3 \\ Layer \\ ReLU} & LinearNet & \shortstack{LinearNet \\ no \\ conv} & CLinear & NLinear & DLinear & \shortstack{TSMixer \\ 1} & \shortstack{TSMixer \\ 2} & \shortstack{TSMixer \\ no \\ relu \\ 1} & \shortstack{TSMixer \\ no \\ relu \\ 2} \\
\shortstack{n \\ ahead} & \shortstack{sequence \\ length} &  &  &  &  &  &  &  &  &  &  &  &  &  &  &  &  &  &  \\
\midrule
\multirow[t]{7}{*}{1} & 3 & \textcolor{blue}{0.0} & \textcolor{blue}{\textbf{\underline{0.0}}} & \textcolor{blue}{7.3} & \textcolor{blue}{7.9} & \textcolor{blue}{8.2} & \textcolor{blue}{8.0} & 4.8 & \textcolor{blue}{4.2} & 5.4 & \textcolor{blue}{7.8} & \textcolor{blue}{7.5} & \textcolor{blue}{4.5} & 6.8 & \textcolor{blue}{6.1} & 16.1 & 14.4 & 16.0 & 16.5 \\
 & 10 & 0.3 & \underline{0.2} & 9.6 & 8.7 & 11.2 & 11.8 & \textcolor{blue}{4.2} & 6.0 & \textcolor{blue}{5.2} & 7.8 & 8.4 & 7.6 & \textcolor{blue}{5.7} & 9.2 & \textcolor{blue}{15.1} & 14.0 & \textcolor{blue}{15.3} & 14.0 \\
 & 20 & 1.1 & \underline{0.2} & 10.0 & 10.4 & 18.2 & 15.8 & 5.5 & 5.9 & 6.0 & 8.2 & 8.9 & 12.3 & 5.7 & 11.8 & 18.4 & 16.3 & 18.4 & \textcolor{blue}{11.5} \\
 & 50 & 4.5 & \underline{0.4} & 13.0 & 16.6 & 29.8 & 29.5 & 5.1 & 6.0 & 6.1 & 11.3 & 10.1 & 12.1 & 7.9 & 8.1 & 20.3 & 15.5 & 21.8 & 17.0 \\
 & 100 & 14.9 & \underline{1.1} & 21.4 & 25.5 & 51.4 & 48.1 & 6.6 & 6.9 & 5.6 & 10.0 & 11.1 & 5.7 & 10.2 & 9.3 & 23.1 & 16.1 & 17.1 & 15.8 \\
 & 200 & nan & nan & 31.3 & 41.5 & 85.2 & 111.0 & 8.1 & 8.4 & \underline{6.0} & 12.5 & 12.7 & 11.4 & 9.4 & 9.3 & 18.5 & \textcolor{blue}{12.7} & 17.6 & 14.3 \\
\bottomrule
\end{tabular}
\end{table}

\subsection{Number of parameters}

\begin{table}[H][H]
\caption{Univariate without Time: Number of Parameters}
\label{tab:univariate_no_time_num_params_btc}
\scriptsize  
\setlength{\tabcolsep}{1pt}  
\begin{tabular}{llllllllllllllllllll}
\toprule
 &  & \shortstack{Linear \\ Regression} & Ridge & \shortstack{LSTM \\ Encoder \\ Decoder} & \shortstack{LSTM \\ Shared \\ Encoder \\ Decoder} & \shortstack{GRU \\ Last \\ Sequence} & \shortstack{GRU \\ Full \\ Sequence} & \shortstack{MLP \\ 1 \\ Layer \\ linear} & \shortstack{MLP \\ 2 \\ Layer \\ linear} & \shortstack{MLP \\ 3 \\ Layer \\ ReLU} & LinearNet & \shortstack{LinearNet \\ no \\ conv} & CLinear & NLinear & DLinear & \shortstack{TSMixer \\ 1} & \shortstack{TSMixer \\ 2} & \shortstack{TSMixer \\ no \\ relu \\ 1} & \shortstack{TSMixer \\ no \\ relu \\ 2} \\
\shortstack{n \\ ahead} & \shortstack{sequence \\ length} &  &  &  &  &  &  &  &  &  &  &  &  &  &  &  &  &  &  \\
\midrule
\multirow[t]{7}{*}{1} & 3 & 4 & 4 & 30651 & 14 & 1061 & 1081 & 4 & 101 & 521 & 62 & 18 & 6 & 6 & 10 & 42 & 76 & 76 & 76 \\
 & 10 & 11 & 11 & 30651 & 14 & 1061 & 1151 & 11 & 241 & 661 & 104 & 32 & 13 & 13 & 24 & 161 & 307 & 307 & 307 \\
 & 20 & 21 & 21 & 30651 & 14 & 1061 & 1251 & 21 & 441 & 861 & 164 & 52 & 23 & 23 & 44 & 501 & 977 & 977 & 977 \\
 & 50 & 51 & 51 & 30651 & 14 & 1061 & 1551 & 51 & 1041 & 1461 & 344 & 112 & 53 & 53 & 104 & 2721 & 5387 & 5387 & 5387 \\
 & 100 & 101 & 101 & 30651 & 14 & 1061 & 2051 & 101 & 2041 & 2461 & 644 & 212 & 103 & 103 & 204 & 10421 & 20737 & 20737 & 20737 \\
 & 200 & nan & nan & 30651 & 14 & 1061 & 3051 & 201 & 4041 & 4461 & 1244 & 412 & 203 & 203 & 404 & 40821 & 81437 & 81437 & 81437 \\
\bottomrule
\end{tabular}
\end{table}

\begin{table}[H][H]
\caption{Univariate with Time: Number of Parameters}
\label{tab:univariate_with_time_num_params_btc}
\scriptsize  
\setlength{\tabcolsep}{1pt}  
\begin{tabular}{llllllllllllllllllll}
\toprule
 &  & \shortstack{Linear \\ Regression} & Ridge & \shortstack{LSTM \\ Encoder \\ Decoder} & \shortstack{LSTM \\ Shared \\ Encoder \\ Decoder} & \shortstack{GRU \\ Last \\ Sequence} & \shortstack{GRU \\ Full \\ Sequence} & \shortstack{MLP \\ 1 \\ Layer \\ linear} & \shortstack{MLP \\ 2 \\ Layer \\ linear} & \shortstack{MLP \\ 3 \\ Layer \\ ReLU} & LinearNet & \shortstack{LinearNet \\ no \\ conv} & CLinear & NLinear & DLinear & \shortstack{TSMixer \\ 1} & \shortstack{TSMixer \\ 2} & \shortstack{TSMixer \\ no \\ relu \\ 1} & \shortstack{TSMixer \\ no \\ relu \\ 2} \\
\shortstack{n \\ ahead} & \shortstack{sequence \\ length} &  &  &  &  &  &  &  &  &  &  &  &  &  &  &  &  &  &  \\
\midrule
\multirow[t]{7}{*}{1} & 3 & 10 & 10 & 31051 & 88 & 1121 & 1141 & 10 & 221 & 641 & 160 & 38 & 8 & 8 & 12 & 82 & 150 & 150 & 150 \\
 & 10 & 31 & 31 & 31051 & 88 & 1121 & 1211 & 31 & 641 & 1061 & 202 & 52 & 15 & 15 & 26 & 229 & 437 & 437 & 437 \\
 & 20 & 61 & 61 & 31051 & 88 & 1121 & 1311 & 61 & 1241 & 1661 & 262 & 72 & 25 & 25 & 46 & 609 & 1187 & 1187 & 1187 \\
 & 50 & 151 & 151 & 31051 & 88 & 1121 & 1611 & 151 & 3041 & 3461 & 442 & 132 & 55 & 55 & 106 & 2949 & 5837 & 5837 & 5837 \\
 & 100 & 301 & 301 & 31051 & 88 & 1121 & 2111 & 301 & 6041 & 6461 & 742 & 232 & 105 & 105 & 206 & 10849 & 21587 & 21587 & 21587 \\
 & 200 & nan & nan & 31051 & 88 & 1121 & 3111 & 601 & 12041 & 12461 & 1342 & 432 & 205 & 205 & 406 & 41649 & 83087 & 83087 & 83087 \\
\bottomrule
\end{tabular}

\end{table}

\begin{table}[H][H]
\centering
\caption{Multivariate without Time: Number of Parameters}
\label{tab:multivariate_no_time_num_params_btc}
\scriptsize  
\setlength{\tabcolsep}{1pt}  
\begin{tabular}{llllllllllllllllllll}
\toprule
 &  & \shortstack{Linear \\ Regression} & Ridge & \shortstack{LSTM \\ Encoder \\ Decoder} & \shortstack{LSTM \\ Shared \\ Encoder \\ Decoder} & \shortstack{GRU \\ Last \\ Sequence} & \shortstack{GRU \\ Full \\ Sequence} & \shortstack{MLP \\ 1 \\ Layer \\ linear} & \shortstack{MLP \\ 2 \\ Layer \\ linear} & \shortstack{MLP \\ 3 \\ Layer \\ ReLU} & LinearNet & \shortstack{LinearNet \\ no \\ conv} & CLinear & NLinear & DLinear & \shortstack{TSMixer \\ 1} & \shortstack{TSMixer \\ 2} & \shortstack{TSMixer \\ no \\ relu \\ 1} & \shortstack{TSMixer \\ no \\ relu \\ 2} \\
\shortstack{n \\ ahead} & \shortstack{sequence \\ length} &  &  &  &  &  &  &  &  &  &  &  &  &  &  &  &  &  &  \\
\midrule
\multirow[t]{7}{*}{1} & 3 & 58 & 58 & 34251 & 2984 & 1601 & 1621 & 58 & 1181 & 1601 & 4616 & 630 & 24 & 24 & 28 & 402 & 742 & 742 & 742 \\
 & 10 & 191 & 191 & 34251 & 2984 & 1601 & 1691 & 191 & 3841 & 4261 & 4658 & 644 & 31 & 31 & 42 & 773 & 1477 & 1477 & 1477 \\
 & 20 & 381 & 381 & 34251 & 2984 & 1601 & 1791 & 381 & 7641 & 8061 & 4718 & 664 & 41 & 41 & 62 & 1473 & 2867 & 2867 & 2867 \\
 & 50 & 951 & 951 & 34251 & 2984 & 1601 & 2091 & 951 & 19041 & 19461 & 4898 & 724 & 71 & 71 & 122 & 4773 & 9437 & 9437 & 9437 \\
 & 100 & 1901 & 1901 & 34251 & 2984 & 1601 & 2591 & 1901 & 38041 & 38461 & 5198 & 824 & 121 & 121 & 222 & 14273 & 28387 & 28387 & 28387 \\
 & 200 & nan & nan & 34251 & 2984 & 1601 & 3591 & 3801 & 76041 & 76461 & 5798 & 1024 & 221 & 221 & 422 & 48273 & 96287 & 96287 & 96287 \\
\bottomrule
\end{tabular}
\end{table}

\begin{table}[H][H]
\centering
\caption{Multivariate with Time: Number of Parameters}
\label{tab:multivariate_with_time_num_params_btc}
\scriptsize  
\setlength{\tabcolsep}{1pt}  
\begin{tabular}{llllllllllllllllllll}
\toprule
 &  & \shortstack{Linear \\ Regression} & Ridge & \shortstack{LSTM \\ Encoder \\ Decoder} & \shortstack{LSTM \\ Shared \\ Encoder \\ Decoder} & \shortstack{GRU \\ Last \\ Sequence} & \shortstack{GRU \\ Full \\ Sequence} & \shortstack{MLP \\ 1 \\ Layer \\ linear} & \shortstack{MLP \\ 2 \\ Layer \\ linear} & \shortstack{MLP \\ 3 \\ Layer \\ ReLU} & LinearNet & \shortstack{LinearNet \\ no \\ conv} & CLinear & NLinear & DLinear & \shortstack{TSMixer \\ 1} & \shortstack{TSMixer \\ 2} & \shortstack{TSMixer \\ no \\ relu \\ 1} & \shortstack{TSMixer \\ no \\ relu \\ 2} \\
\shortstack{n \\ ahead} & \shortstack{sequence \\ length} &  &  &  &  &  &  &  &  &  &  &  &  &  &  &  &  &  &  \\
\midrule
\multirow[t]{7}{*}{1} & 3 & 64 & 64 & 34651 & 3634 & 1661 & 1681 & 64 & 1301 & 1721 & 5632 & 758 & 26 & 26 & 30 & 442 & 816 & 816 & 816 \\
 & 10 & 211 & 211 & 34651 & 3634 & 1661 & 1751 & 211 & 4241 & 4661 & 5674 & 772 & 33 & 33 & 44 & 841 & 1607 & 1607 & 1607 \\
 & 20 & 421 & 421 & 34651 & 3634 & 1661 & 1851 & 421 & 8441 & 8861 & 5734 & 792 & 43 & 43 & 64 & 1581 & 3077 & 3077 & 3077 \\
 & 50 & 1051 & 1051 & 34651 & 3634 & 1661 & 2151 & 1051 & 21041 & 21461 & 5914 & 852 & 73 & 73 & 124 & 5001 & 9887 & 9887 & 9887 \\
 & 100 & 2101 & 2101 & 34651 & 3634 & 1661 & 2651 & 2101 & 42041 & 42461 & 6214 & 952 & 123 & 123 & 224 & 14701 & 29237 & 29237 & 29237 \\
 & 200 & nan & nan & 34651 & 3634 & 1661 & 3651 & 4201 & 84041 & 84461 & 6814 & 1152 & 223 & 223 & 424 & 49101 & 97937 & 97937 & 97937 \\
\bottomrule
\end{tabular}
\end{table}

\section{ETTh1 prediction tasks results}
\subsection{MSE}

\begin{table}[H]
\caption{Univariate without Time: Mean Squared Error (x1e2)}
\label{tab:univariate_no_time_mse_etth1}
\scriptsize  
\setlength{\tabcolsep}{2pt}  
\begin{tabular}{llllllllllllllllllll}
\toprule
 &  & \shortstack{Linear \\ Regression} & Ridge & \shortstack{LSTM \\ Encoder \\ Decoder} & \shortstack{LSTM \\ Shared \\ Encoder \\ Decoder} & \shortstack{GRU \\ Last \\ Sequence} & \shortstack{GRU \\ Full \\ Sequence} & \shortstack{MLP \\ 1 \\ Layer \\ linear} & \shortstack{MLP \\ 2 \\ Layer \\ linear} & \shortstack{MLP \\ 3 \\ Layer \\ ReLU} & LinearNet & \shortstack{LinearNet \\ no \\ conv} & CLinear & NLinear & DLinear & \shortstack{TSMixer \\ 1} & \shortstack{TSMixer \\ 2} & \shortstack{TSMixer \\ no \\ relu \\ 1} & \shortstack{TSMixer \\ no \\ relu \\ 2} \\
\shortstack{n \\ ahead} & \shortstack{sequence \\ length} &  &  &  &  &  &  &  &  &  &  &  &  &  &  &  &  &  &  \\
\midrule
\multirow[t]{7}{*}{1} & 3 & 0.87 & 0.87 & 0.91 & 3.40 & 0.93 & 0.90 & 0.88 & 0.88 & 0.88 & 0.88 & 0.88 & 0.89 & 0.87 & 0.87 & 0.87 & \underline{0.86} & 0.93 & 0.87 \\
 & 10 & 0.86 & 0.86 & 0.88 & 1.73 & 0.94 & 0.89 & 0.89 & 0.87 & 0.89 & 0.87 & 0.87 & 0.89 & 0.87 & 0.88 & \underline{0.85} & 0.86 & 0.86 & 0.86 \\
 & 20 & 0.84 & \underline{0.84} & 0.86 & 1.56 & \textcolor{blue}{0.88} & 0.86 & 0.85 & 0.85 & \textcolor{blue}{0.87} & 0.85 & 0.86 & 0.84 & 0.84 & 0.86 & 93.89 & \textcolor{blue}{0.84} & 131.75 & 0.85 \\
 & 50 & 0.81 & 0.81 & 0.87 & \textcolor{blue}{1.40} & 0.91 & 0.85 & 0.82 & 0.82 & 0.88 & 0.84 & 0.83 & 0.84 & \textcolor{blue}{0.81} & 0.81 & \textcolor{blue}{0.83} & 92.80 & 106.00 & \textcolor{blue}{\underline{0.80}} \\
 & 100 & \textcolor{blue}{\underline{0.79}} & \textcolor{blue}{0.79} & 0.85 & 3.96 & 0.88 & 0.82 & 0.83 & \textcolor{blue}{0.81} & 0.88 & 0.81 & 0.82 & 0.82 & 135.52 & 0.79 & 121.46 & 128.87 & 0.79 & 0.80 \\
 & 200 & nan & nan & 0.86 & 1.78 & 0.93 & \textcolor{blue}{\underline{0.79}} & \textcolor{blue}{0.82} & 0.82 & 1.00 & \textcolor{blue}{0.80} & 0.82 & \textcolor{blue}{0.81} & 130.67 & 0.79 & 175.63 & 1.16 & 0.79 & 0.84 \\
\bottomrule
\end{tabular}
\end{table}

\begin{table}[H]
\caption{Univariate with Time: Mean Squared Error}
\label{tab:univariate_with_time_mse_etth1}
\scriptsize  
\setlength{\tabcolsep}{2pt}  
\begin{tabular}{llllllllllllllllllll}
\toprule
 &  & \shortstack{Linear \\ Regression} & Ridge & \shortstack{LSTM \\ Encoder \\ Decoder} & \shortstack{LSTM \\ Shared \\ Encoder \\ Decoder} & \shortstack{GRU \\ Last \\ Sequence} & \shortstack{GRU \\ Full \\ Sequence} & \shortstack{MLP \\ 1 \\ Layer \\ linear} & \shortstack{MLP \\ 2 \\ Layer \\ linear} & \shortstack{MLP \\ 3 \\ Layer \\ ReLU} & LinearNet & \shortstack{LinearNet \\ no \\ conv} & CLinear & NLinear & DLinear & \shortstack{TSMixer \\ 1} & \shortstack{TSMixer \\ 2} & \shortstack{TSMixer \\ no \\ relu \\ 1} & \shortstack{TSMixer \\ no \\ relu \\ 2} \\
\shortstack{n \\ ahead} & \shortstack{sequence \\ length} &  &  &  &  &  &  &  &  &  &  &  &  &  &  &  &  &  &  \\
\midrule
\multirow[t]{7}{*}{1} & 3 & 0.87 & 0.87 & \underline{0.86} & 2.13 & 0.93 & 0.90 & 0.87 & 0.87 & 0.88 & 0.89 & 0.88 & 0.87 & 0.87 & 0.87 & 0.91 & 0.87 & 0.87 & 0.87 \\
 & 10 & 0.86 & 0.86 & 0.89 & 5.47 & 0.97 & 0.94 & 0.88 & 0.86 & 0.87 & 0.90 & 0.87 & 0.88 & 0.86 & 0.86 & \underline{0.86} & \textcolor{blue}{0.87} & 0.86 & 0.87 \\
 & 20 & 0.84 & \underline{0.84} & 0.87 & 1.83 & 0.88 & 0.85 & 0.85 & 0.85 & \textcolor{blue}{0.85} & 0.85 & 0.86 & 0.85 & 0.84 & 0.84 & \textcolor{blue}{0.85} & 111.00 & 131.07 & 133.02 \\
 & 50 & \underline{0.81} & 0.81 & 0.87 & 1.40 & 0.89 & 0.85 & 0.83 & 0.83 & 0.88 & 0.83 & 0.84 & 0.84 & \textcolor{blue}{0.82} & 0.81 & 75.17 & 98.08 & 109.19 & 89.21 \\
 & 100 & \textcolor{blue}{\textbf{\underline{0.79}}} & \textcolor{blue}{0.79} & \textcolor{blue}{0.83} & 1.93 & \textcolor{blue}{0.87} & 0.87 & \textcolor{blue}{0.83} & \textcolor{blue}{0.80} & 0.87 & 0.82 & 0.82 & \textcolor{blue}{0.82} & 135.43 & 0.80 & 112.55 & 0.92 & \textcolor{blue}{0.79} & 124.02 \\
 & 200 & nan & nan & 0.83 & \textcolor{blue}{1.32} & 0.90 & \textcolor{blue}{0.79} & 0.83 & 0.81 & 0.94 & 0.80 & \textcolor{blue}{0.81} & 0.82 & 132.47 & \underline{0.79} & 462.56 & 121.72 & 0.85 & \textcolor{blue}{0.79} \\
\bottomrule
\end{tabular}
\end{table}

\begin{table}[H]
\caption{Multivariate without Time: Mean Squared Error}
\label{tab:multivariate_no_time_mse_etth1}
\scriptsize  
\setlength{\tabcolsep}{2pt}  
\begin{tabular}{llllllllllllllllllll}
\toprule
 &  & \shortstack{Linear \\ Regression} & Ridge & \shortstack{LSTM \\ Encoder \\ Decoder} & \shortstack{LSTM \\ Shared \\ Encoder \\ Decoder} & \shortstack{GRU \\ Last \\ Sequence} & \shortstack{GRU \\ Full \\ Sequence} & \shortstack{MLP \\ 1 \\ Layer \\ linear} & \shortstack{MLP \\ 2 \\ Layer \\ linear} & \shortstack{MLP \\ 3 \\ Layer \\ ReLU} & LinearNet & \shortstack{LinearNet \\ no \\ conv} & CLinear & NLinear & DLinear & \shortstack{TSMixer \\ 1} & \shortstack{TSMixer \\ 2} & \shortstack{TSMixer \\ no \\ relu \\ 1} & \shortstack{TSMixer \\ no \\ relu \\ 2} \\
\shortstack{n \\ ahead} & \shortstack{sequence \\ length} &  &  &  &  &  &  &  &  &  &  &  &  &  &  &  &  &  &  \\
\midrule
\multirow[t]{7}{*}{1} & 3 & 0.87 & 0.87 & 1.09 & 1.62 & 1.38 & 1.59 & 1.66 & 0.95 & \textcolor{blue}{1.34} & \underline{0.87} & 0.95 & 1.04 & 1.10 & 0.92 & \textcolor{blue}{0.93} & 4.16 & 1.58 & 1.91 \\
 & 10 & \underline{0.84} & 0.84 & 1.24 & 1.09 & \textcolor{blue}{1.03} & 1.15 & \textcolor{blue}{1.22} & 1.13 & 1.38 & 0.89 & 0.98 & 0.95 & 1.26 & 0.96 & 0.99 & \textcolor{blue}{3.59} & 2.34 & \textcolor{blue}{0.95} \\
 & 20 & 0.81 & \underline{0.81} & 1.23 & \textcolor{blue}{1.04} & 1.07 & \textcolor{blue}{0.99} & 2.94 & 2.14 & 1.98 & 0.95 & 0.96 & \textcolor{blue}{0.86} & 1.92 & 1.54 & 2.74 & 228.76 & 0.93 & 1.52 \\
 & 50 & 0.80 & \underline{0.79} & 1.43 & 2.08 & 1.39 & 1.20 & 3.86 & \textcolor{blue}{0.88} & 5.24 & 0.99 & 0.84 & 0.86 & 0.91 & 0.91 & 1.15 & 4.62 & \textcolor{blue}{0.83} & 322.19 \\
 & 100 & \textcolor{blue}{0.79} & \textcolor{blue}{\textbf{\underline{0.78}}} & 1.14 & 1.38 & 1.38 & 1.08 & 1.78 & 2.85 & 7.61 & 0.98 & 0.87 & 1.53 & \textcolor{blue}{0.86} & \textcolor{blue}{0.86} & 14.06 & 8.12 & 124.68 & 0.99 \\
 & 200 & nan & nan & \textcolor{blue}{1.05} & 12.48 & 1.06 & 1.26 & 1.52 & 2.96 & 17.34 & \textcolor{blue}{0.82} & \textcolor{blue}{\underline{0.81}} & 272.42 & 1.91 & 4.06 & 27.81 & 246.04 & 331.66 & 121.42 \\
\bottomrule
\end{tabular}

\end{table}

\begin{table}[H]
\caption{Multivariate with Time: Mean Squared Error}
\label{tab:multivariate_with_time_mse_etth1}
\scriptsize  
\setlength{\tabcolsep}{2pt}  
\begin{tabular}{llllllllllllllllllll}
\toprule
 &  & \shortstack{Linear \\ Regression} & Ridge & \shortstack{LSTM \\ Encoder \\ Decoder} & \shortstack{LSTM \\ Shared \\ Encoder \\ Decoder} & \shortstack{GRU \\ Last \\ Sequence} & \shortstack{GRU \\ Full \\ Sequence} & \shortstack{MLP \\ 1 \\ Layer \\ linear} & \shortstack{MLP \\ 2 \\ Layer \\ linear} & \shortstack{MLP \\ 3 \\ Layer \\ ReLU} & LinearNet & \shortstack{LinearNet \\ no \\ conv} & CLinear & NLinear & DLinear & \shortstack{TSMixer \\ 1} & \shortstack{TSMixer \\ 2} & \shortstack{TSMixer \\ no \\ relu \\ 1} & \shortstack{TSMixer \\ no \\ relu \\ 2} \\
\shortstack{n \\ ahead} & \shortstack{sequence \\ length} &  &  &  &  &  &  &  &  &  &  &  &  &  &  &  &  &  &  \\
\midrule
\multirow[t]{7}{*}{1} & 3 & \underline{0.86} & 0.86 & \textcolor{blue}{1.05} & 1.27 & 1.66 & \textcolor{blue}{1.13} & \textcolor{blue}{0.92} & 1.00 & \textcolor{blue}{1.17} & 0.87 & 0.94 & 0.88 & 0.88 & 1.22 & \textcolor{blue}{0.91} & 1.03 & 2.35 & 1.84 \\
 & 10 & \underline{0.82} & 0.82 & 1.28 & 1.83 & \textcolor{blue}{1.11} & 1.35 & 0.98 & 1.20 & 1.54 & 0.89 & 0.96 & 1.13 & 1.09 & \textcolor{blue}{1.19} & 0.97 & 0.92 & 1.62 & 0.98 \\
 & 20 & 0.80 & \underline{0.80} & 1.11 & 1.29 & 1.27 & 1.16 & 1.28 & \textcolor{blue}{0.89} & 2.39 & 0.86 & 0.94 & \textcolor{blue}{0.87} & 0.88 & 2.01 & 234.07 & \textcolor{blue}{0.91} & \textcolor{blue}{0.91} & 1.08 \\
 & 50 & 0.79 & \underline{0.79} & 1.88 & 1.71 & 3.24 & 1.25 & 1.16 & 1.22 & 3.90 & 0.85 & 0.93 & 0.96 & 0.93 & 2.21 & 1.02 & 0.97 & 1.02 & \textcolor{blue}{0.86} \\
 & 100 & \textcolor{blue}{0.79} & \textcolor{blue}{\textbf{\underline{0.79}}} & 1.62 & 2.30 & 1.31 & 2.04 & 1.36 & 1.04 & 9.77 & 0.85 & 1.45 & 1.39 & \textcolor{blue}{0.80} & 2.96 & 147.32 & 3.10 & 1.62 & 185.25 \\
 & 200 & nan & nan & 1.49 & \textcolor{blue}{1.21} & 1.48 & 2.35 & 1.47 & 2.62 & 18.57 & \textcolor{blue}{\underline{0.82}} & 1.39 & 1.09 & 1.34 & 5.32 & 393.69 & 151.92 & 1.29 & 4.69 \\
\bottomrule
\end{tabular}
\end{table}

\subsection{Training Times}

\begin{table}[H]
\caption{Univariate without Time: Training time}
\label{tab:univariate_no_time_train_time_etth1}
\scriptsize  
\setlength{\tabcolsep}{2pt}  
\begin{tabular}{llllllllllllllllllll}
\toprule
 &  & \shortstack{Linear \\ Regression} & Ridge & \shortstack{LSTM \\ Encoder \\ Decoder} & \shortstack{LSTM \\ Shared \\ Encoder \\ Decoder} & \shortstack{GRU \\ Last \\ Sequence} & \shortstack{GRU \\ Full \\ Sequence} & \shortstack{MLP \\ 1 \\ Layer \\ linear} & \shortstack{MLP \\ 2 \\ Layer \\ linear} & \shortstack{MLP \\ 3 \\ Layer \\ ReLU} & LinearNet & \shortstack{LinearNet \\ no \\ conv} & CLinear & NLinear & DLinear & \shortstack{TSMixer \\ 1} & \shortstack{TSMixer \\ 2} & \shortstack{TSMixer \\ no \\ relu \\ 1} & \shortstack{TSMixer \\ no \\ relu \\ 2} \\
\shortstack{n \\ ahead} & \shortstack{sequence \\ length} &  &  &  &  &  &  &  &  &  &  &  &  &  &  &  &  &  &  \\
\midrule
\multirow[t]{7}{*}{1} & 3 & \textcolor{blue}{\textbf{\underline{0.0}}} & \textcolor{blue}{0.0} & \textcolor{blue}{6.7} & \textcolor{blue}{7.5} & \textcolor{blue}{7.7} & \textcolor{blue}{7.7} & 9.5 & 5.2 & \textcolor{blue}{5.5} & 9.3 & 10.9 & 8.0 & 5.9 & 7.4 & 10.9 & 9.2 & 9.5 & 9.5 \\
 & 10 & 0.0 & \underline{0.0} & 7.9 & 16.3 & 9.5 & 12.0 & 8.5 & \textcolor{blue}{3.5} & 7.2 & 10.5 & 10.1 & \textcolor{blue}{6.9} & 7.8 & 4.4 & 12.2 & 12.7 & 12.3 & 15.5 \\
 & 20 & 0.0 & \underline{0.0} & 9.4 & 22.6 & 16.9 & 15.0 & 9.7 & 4.3 & 7.5 & \textcolor{blue}{9.2} & 10.4 & 13.3 & 8.7 & 3.6 & 7.8 & 11.8 & \textcolor{blue}{8.1} & 10.2 \\
 & 50 & 0.0 & \underline{0.0} & 14.6 & 20.7 & 25.0 & 32.0 & 7.6 & 5.3 & 7.9 & 10.2 & 9.8 & 8.8 & 8.3 & 4.4 & 10.8 & 16.3 & 15.1 & 15.6 \\
 & 100 & 0.0 & \underline{0.0} & 21.9 & 14.4 & 35.0 & 44.9 & \textcolor{blue}{7.2} & 7.5 & 8.6 & 10.6 & \textcolor{blue}{9.8} & 9.7 & \textcolor{blue}{4.1} & \textcolor{blue}{3.5} & 6.3 & 15.1 & 11.3 & 9.7 \\
 & 200 & nan & nan & 37.0 & 60.2 & 82.5 & 85.4 & 9.1 & 6.8 & 7.6 & 10.6 & 11.7 & 9.1 & 7.2 & \underline{4.6} & \textcolor{blue}{6.2} & \textcolor{blue}{8.5} & 9.7 & 10.1 \\
\bottomrule
\end{tabular}

\end{table}

\begin{table}[H]
\caption{Univariate with Time: Training time}
\label{tab:univariate_with_time_train_time_etth1}
\scriptsize  
\setlength{\tabcolsep}{2pt}  
\begin{tabular}{llllllllllllllllllll}
\toprule
 &  & \shortstack{Linear \\ Regression} & Ridge & \shortstack{LSTM \\ Encoder \\ Decoder} & \shortstack{LSTM \\ Shared \\ Encoder \\ Decoder} & \shortstack{GRU \\ Last \\ Sequence} & \shortstack{GRU \\ Full \\ Sequence} & \shortstack{MLP \\ 1 \\ Layer \\ linear} & \shortstack{MLP \\ 2 \\ Layer \\ linear} & \shortstack{MLP \\ 3 \\ Layer \\ ReLU} & LinearNet & \shortstack{LinearNet \\ no \\ conv} & CLinear & NLinear & DLinear & \shortstack{TSMixer \\ 1} & \shortstack{TSMixer \\ 2} & \shortstack{TSMixer \\ no \\ relu \\ 1} & \shortstack{TSMixer \\ no \\ relu \\ 2} \\
\shortstack{n \\ ahead} & \shortstack{sequence \\ length} &  &  &  &  &  &  &  &  &  &  &  &  &  &  &  &  &  &  \\
\midrule
\multirow[t]{7}{*}{1} & 3 & \textcolor{blue}{0.0} & \textcolor{blue}{\textbf{\underline{0.0}}} & \textcolor{blue}{7.1} & 8.5 & \textcolor{blue}{7.2} & \textcolor{blue}{7.2} & \textcolor{blue}{7.7} & \textcolor{blue}{3.3} & \textcolor{blue}{4.6} & \textcolor{blue}{9.2} & \textcolor{blue}{5.2} & \textcolor{blue}{4.1} & 5.1 & 3.8 & 8.6 & 10.6 & 12.2 & 7.9 \\
 & 10 & 0.0 & \underline{0.0} & 9.0 & \textcolor{blue}{6.4} & 8.5 & 9.9 & 9.5 & 3.9 & 5.3 & 9.4 & 9.3 & 11.7 & 7.8 & 4.6 & 9.6 & 15.2 & 13.8 & 9.5 \\
 & 20 & 0.0 & \underline{0.0} & 9.4 & 41.1 & 13.9 & 15.3 & 9.5 & 5.5 & 7.7 & 10.5 & 9.7 & 11.4 & 4.8 & \textcolor{blue}{3.1} & 7.9 & \textcolor{blue}{8.4} & \textcolor{blue}{8.0} & \textcolor{blue}{7.7} \\
 & 50 & 0.0 & \underline{0.0} & 19.1 & 35.0 & 30.6 & 33.0 & 9.2 & 5.5 & 6.7 & 11.0 & 9.8 & 8.8 & 4.6 & 4.6 & 9.5 & 14.6 & 14.5 & 15.7 \\
 & 100 & 0.0 & \underline{0.0} & 28.8 & 52.3 & 44.9 & 46.8 & 7.8 & 7.6 & 7.9 & 10.2 & 10.8 & 9.1 & 5.3 & 6.3 & \textcolor{blue}{6.2} & 9.6 & 9.2 & 9.3 \\
 & 200 & nan & nan & 44.7 & 60.0 & 70.9 & 89.5 & 9.7 & 9.7 & 8.0 & 11.1 & 13.6 & 10.0 & 5.0 & \underline{4.3} & 6.6 & 8.9 & 9.3 & 10.5 \\
\bottomrule
\end{tabular}
\end{table}

\begin{table}[H]
\caption{Multivariate without Time: Training time}
\label{tab:multivariate_no_time_train_time_etth1}
\scriptsize  
\setlength{\tabcolsep}{2pt}  
\begin{tabular}{llllllllllllllllllll}
\toprule
 &  & \shortstack{Linear \\ Regression} & Ridge & \shortstack{LSTM \\ Encoder \\ Decoder} & \shortstack{LSTM \\ Shared \\ Encoder \\ Decoder} & \shortstack{GRU \\ Last \\ Sequence} & \shortstack{GRU \\ Full \\ Sequence} & \shortstack{MLP \\ 1 \\ Layer \\ linear} & \shortstack{MLP \\ 2 \\ Layer \\ linear} & \shortstack{MLP \\ 3 \\ Layer \\ ReLU} & LinearNet & \shortstack{LinearNet \\ no \\ conv} & CLinear & NLinear & DLinear & \shortstack{TSMixer \\ 1} & \shortstack{TSMixer \\ 2} & \shortstack{TSMixer \\ no \\ relu \\ 1} & \shortstack{TSMixer \\ no \\ relu \\ 2} \\
\shortstack{n \\ ahead} & \shortstack{sequence \\ length} &  &  &  &  &  &  &  &  &  &  &  &  &  &  &  &  &  &  \\
\midrule
\multirow[t]{7}{*}{1} & 3 & \textcolor{blue}{0.0} & \textcolor{blue}{\textbf{\underline{0.0}}} & 7.6 & \textcolor{blue}{8.1} & \textcolor{blue}{8.0} & 49.4 & 4.4 & 4.3 & 6.0 & \textcolor{blue}{7.7} & 8.0 & 22.2 & 9.4 & 5.8 & \textcolor{blue}{13.6} & 13.5 & 15.4 & 17.8 \\
 & 10 & 0.0 & \underline{0.0} & \textcolor{blue}{7.4} & 12.9 & 14.1 & \textcolor{blue}{13.5} & 4.6 & 4.6 & 6.4 & 9.1 & \textcolor{blue}{7.9} & 11.0 & 40.1 & 14.4 & 14.8 & 11.5 & 12.4 & 13.7 \\
 & 20 & 0.1 & \underline{0.1} & 7.6 & 15.2 & 19.6 & 22.1 & 4.5 & 3.8 & 7.5 & 7.9 & 11.0 & 11.4 & 60.6 & 10.6 & 16.5 & \textcolor{blue}{9.8} & 21.3 & 14.1 \\
 & 50 & 0.6 & \underline{0.1} & 9.8 & 11.7 & 43.6 & 31.4 & 4.3 & 6.4 & 5.5 & 12.1 & 12.1 & 9.2 & 7.7 & 6.0 & 20.8 & 19.8 & 22.7 & 9.4 \\
 & 100 & 1.7 & \underline{0.3} & 12.9 & 30.2 & 42.2 & 60.5 & \textcolor{blue}{4.0} & 3.4 & 4.6 & 8.3 & 9.2 & 8.5 & 7.7 & 7.3 & 23.8 & 15.7 & 13.2 & 19.2 \\
 & 200 & nan & nan & 28.1 & 14.9 & 118.9 & 107.9 & 5.4 & 3.4 & 4.8 & 11.3 & 12.4 & \textcolor{blue}{\underline{2.8}} & \textcolor{blue}{7.0} & \textcolor{blue}{3.7} & 14.1 & 10.1 & 12.0 & 11.3 \\
\bottomrule
\end{tabular}

\end{table}

\begin{table}[H]
\caption{Multivariate with Time: Training time}
\label{tab:multivariate_with_time_train_time_etth1}
\scriptsize  
\setlength{\tabcolsep}{2pt}  
\begin{tabular}{llllllllllllllllllll}
\toprule
 &  & \shortstack{Linear \\ Regression} & Ridge & \shortstack{LSTM \\ Encoder \\ Decoder} & \shortstack{LSTM \\ Shared \\ Encoder \\ Decoder} & \shortstack{GRU \\ Last \\ Sequence} & \shortstack{GRU \\ Full \\ Sequence} & \shortstack{MLP \\ 1 \\ Layer \\ linear} & \shortstack{MLP \\ 2 \\ Layer \\ linear} & \shortstack{MLP \\ 3 \\ Layer \\ ReLU} & LinearNet & \shortstack{LinearNet \\ no \\ conv} & CLinear & NLinear & DLinear & \shortstack{TSMixer \\ 1} & \shortstack{TSMixer \\ 2} & \shortstack{TSMixer \\ no \\ relu \\ 1} & \shortstack{TSMixer \\ no \\ relu \\ 2} \\
\shortstack{n \\ ahead} & \shortstack{sequence \\ length} &  &  &  &  &  &  &  &  &  &  &  &  &  &  &  &  &  &  \\
\midrule
\multirow[t]{7}{*}{1} & 3 & \textcolor{blue}{0.0} & \textcolor{blue}{\textbf{\underline{0.0}}} & \textcolor{blue}{6.9} & 10.1 & \textcolor{blue}{9.1} & \textcolor{blue}{6.9} & 6.7 & \textcolor{blue}{3.7} & 5.5 & 8.4 & 7.5 & 11.6 & \textcolor{blue}{6.7} & 3.7 & 14.5 & 16.3 & 56.2 & 12.8 \\
 & 10 & 0.0 & \underline{0.0} & 7.4 & \textcolor{blue}{8.5} & 12.8 & 14.0 & 5.2 & 3.8 & 6.6 & 11.3 & \textcolor{blue}{6.1} & 4.4 & 20.1 & 11.0 & 15.8 & 13.7 & 16.0 & 12.8 \\
 & 20 & 0.1 & \underline{0.1} & 10.3 & 9.2 & 13.1 & 16.0 & \textcolor{blue}{4.6} & 4.4 & 6.0 & 10.2 & 9.0 & 11.5 & 9.9 & 9.2 & \textcolor{blue}{11.6} & 17.6 & 17.7 & 17.4 \\
 & 50 & 0.8 & \underline{0.2} & 17.1 & 19.6 & 23.2 & 36.3 & 5.6 & 4.0 & 5.5 & 10.8 & 13.4 & 10.7 & 10.4 & 7.5 & 17.5 & 16.4 & 16.7 & 18.2 \\
 & 100 & 3.0 & \underline{0.4} & 18.5 & 18.5 & 68.7 & 130.4 & 5.5 & 5.3 & 4.9 & \textcolor{blue}{8.2} & 8.5 & 8.3 & 10.9 & 4.9 & 17.8 & 18.1 & 17.3 & 11.9 \\
 & 200 & nan & nan & 27.1 & 55.5 & 91.9 & 49.7 & 7.2 & 4.2 & \textcolor{blue}{3.7} & 13.6 & 10.1 & 14.1 & 7.3 & \textcolor{blue}{\underline{3.6}} & 12.7 & \textcolor{blue}{11.5} & 20.1 & 40.9 \\
\bottomrule
\end{tabular}

\end{table}

\subsection{Number of parameters}

\begin{table}[H]
\caption{Univariate without Time: Number of Parameters}
\label{tab:univariate_no_time_num_params_etth1}
\scriptsize  
\setlength{\tabcolsep}{1pt}  
\begin{tabular}{llllllllllllllllllll}
\toprule
 &  & \shortstack{Linear \\ Regression} & Ridge & \shortstack{LSTM \\ Encoder \\ Decoder} & \shortstack{LSTM \\ Shared \\ Encoder \\ Decoder} & \shortstack{GRU \\ Last \\ Sequence} & \shortstack{GRU \\ Full \\ Sequence} & \shortstack{MLP \\ 1 \\ Layer \\ linear} & \shortstack{MLP \\ 2 \\ Layer \\ linear} & \shortstack{MLP \\ 3 \\ Layer \\ ReLU} & LinearNet & \shortstack{LinearNet \\ no \\ conv} & CLinear & NLinear & DLinear & \shortstack{TSMixer \\ 1} & \shortstack{TSMixer \\ 2} & \shortstack{TSMixer \\ no \\ relu \\ 1} & \shortstack{TSMixer \\ no \\ relu \\ 2} \\
\shortstack{n \\ ahead} & \shortstack{sequence \\ length} &  &  &  &  &  &  &  &  &  &  &  &  &  &  &  &  &  &  \\
\midrule
\multirow[t]{7}{*}{1} & 3 & 4 & 4 & 30651 & 14 & 1061 & 1081 & 4 & 101 & 521 & 62 & 18 & 6 & 6 & 10 & 42 & 76 & 76 & 76 \\
 & 10 & 11 & 11 & 30651 & 14 & 1061 & 1151 & 11 & 241 & 661 & 104 & 32 & 13 & 13 & 24 & 161 & 307 & 307 & 307 \\
 & 20 & 21 & 21 & 30651 & 14 & 1061 & 1251 & 21 & 441 & 861 & 164 & 52 & 23 & 23 & 44 & 501 & 977 & 977 & 977 \\
 & 50 & 51 & 51 & 30651 & 14 & 1061 & 1551 & 51 & 1041 & 1461 & 344 & 112 & 53 & 53 & 104 & 2721 & 5387 & 5387 & 5387 \\
 & 100 & 101 & 101 & 30651 & 14 & 1061 & 2051 & 101 & 2041 & 2461 & 644 & 212 & 103 & 103 & 204 & 10421 & 20737 & 20737 & 20737 \\
 & 200 & nan & nan & 30651 & 14 & 1061 & 3051 & 201 & 4041 & 4461 & 1244 & 412 & 203 & 203 & 404 & 40821 & 81437 & 81437 & 81437 \\
\bottomrule
\end{tabular}
\end{table}

\begin{table}[H]
\caption{Univariate with Time: Number of Parameters}
\label{tab:univariate_with_time_num_params_etth1}
\scriptsize  
\setlength{\tabcolsep}{1pt}  
\begin{tabular}{llllllllllllllllllll}
\toprule
 &  & \shortstack{Linear \\ Regression} & Ridge & \shortstack{LSTM \\ Encoder \\ Decoder} & \shortstack{LSTM \\ Shared \\ Encoder \\ Decoder} & \shortstack{GRU \\ Last \\ Sequence} & \shortstack{GRU \\ Full \\ Sequence} & \shortstack{MLP \\ 1 \\ Layer \\ linear} & \shortstack{MLP \\ 2 \\ Layer \\ linear} & \shortstack{MLP \\ 3 \\ Layer \\ ReLU} & LinearNet & \shortstack{LinearNet \\ no \\ conv} & CLinear & NLinear & DLinear & \shortstack{TSMixer \\ 1} & \shortstack{TSMixer \\ 2} & \shortstack{TSMixer \\ no \\ relu \\ 1} & \shortstack{TSMixer \\ no \\ relu \\ 2} \\
\shortstack{n \\ ahead} & \shortstack{sequence \\ length} &  &  &  &  &  &  &  &  &  &  &  &  &  &  &  &  &  &  \\
\midrule
\multirow[t]{7}{*}{1} & 3 & 4 & 4 & 30651 & 14 & 1061 & 1081 & 4 & 101 & 521 & 62 & 18 & 6 & 6 & 10 & 42 & 76 & 76 & 76 \\
 & 10 & 11 & 11 & 30651 & 14 & 1061 & 1151 & 11 & 241 & 661 & 104 & 32 & 13 & 13 & 24 & 161 & 307 & 307 & 307 \\
 & 20 & 21 & 21 & 30651 & 14 & 1061 & 1251 & 21 & 441 & 861 & 164 & 52 & 23 & 23 & 44 & 501 & 977 & 977 & 977 \\
 & 50 & 51 & 51 & 30651 & 14 & 1061 & 1551 & 51 & 1041 & 1461 & 344 & 112 & 53 & 53 & 104 & 2721 & 5387 & 5387 & 5387 \\
 & 100 & 101 & 101 & 30651 & 14 & 1061 & 2051 & 101 & 2041 & 2461 & 644 & 212 & 103 & 103 & 204 & 10421 & 20737 & 20737 & 20737 \\
 & 200 & nan & nan & 30651 & 14 & 1061 & 3051 & 201 & 4041 & 4461 & 1244 & 412 & 203 & 203 & 404 & 40821 & 81437 & 81437 & 81437 \\
\bottomrule
\end{tabular}

\end{table}

\begin{table}[H]
\caption{Multivariate without Time: Number of Parameters}
\label{tab:multivariate_no_time_num_params_etth1}
\scriptsize  
\setlength{\tabcolsep}{1pt}  
\begin{tabular}{llllllllllllllllllll}
\toprule
 &  & \shortstack{Linear \\ Regression} & Ridge & \shortstack{LSTM \\ Encoder \\ Decoder} & \shortstack{LSTM \\ Shared \\ Encoder \\ Decoder} & \shortstack{GRU \\ Last \\ Sequence} & \shortstack{GRU \\ Full \\ Sequence} & \shortstack{MLP \\ 1 \\ Layer \\ linear} & \shortstack{MLP \\ 2 \\ Layer \\ linear} & \shortstack{MLP \\ 3 \\ Layer \\ ReLU} & LinearNet & \shortstack{LinearNet \\ no \\ conv} & CLinear & NLinear & DLinear & \shortstack{TSMixer \\ 1} & \shortstack{TSMixer \\ 2} & \shortstack{TSMixer \\ no \\ relu \\ 1} & \shortstack{TSMixer \\ no \\ relu \\ 2} \\
\shortstack{n \\ ahead} & \shortstack{sequence \\ length} &  &  &  &  &  &  &  &  &  &  &  &  &  &  &  &  &  &  \\
\midrule
\multirow[t]{7}{*}{1} & 3 & 22 & 22 & 31851 & 428 & 1241 & 1261 & 22 & 461 & 881 & 662 & 114 & 12 & 12 & 16 & 162 & 298 & 298 & 298 \\
 & 10 & 71 & 71 & 31851 & 428 & 1241 & 1331 & 71 & 1441 & 1861 & 704 & 128 & 19 & 19 & 30 & 365 & 697 & 697 & 697 \\
 & 20 & 141 & 141 & 31851 & 428 & 1241 & 1431 & 141 & 2841 & 3261 & 764 & 148 & 29 & 29 & 50 & 825 & 1607 & 1607 & 1607 \\
 & 50 & 351 & 351 & 31851 & 428 & 1241 & 1731 & 351 & 7041 & 7461 & 944 & 208 & 59 & 59 & 110 & 3405 & 6737 & 6737 & 6737 \\
 & 100 & 701 & 701 & 31851 & 428 & 1241 & 2231 & 701 & 14041 & 14461 & 1244 & 308 & 109 & 109 & 210 & 11705 & 23287 & 23287 & 23287 \\
 & 200 & nan & nan & 31851 & 428 & 1241 & 3231 & 1401 & 28041 & 28461 & 1844 & 508 & 209 & 209 & 410 & 43305 & 86387 & 86387 & 86387 \\
\bottomrule
\end{tabular}

\end{table}

\begin{table}[H]
\caption{Multivariate with Time: Number of Parameters}
\label{tab:multivariate_with_time_num_params_etth1}
\scriptsize  
\setlength{\tabcolsep}{1pt}  
\begin{tabular}{llllllllllllllllllll}
\toprule
 &  & \shortstack{Linear \\ Regression} & Ridge & \shortstack{LSTM \\ Encoder \\ Decoder} & \shortstack{LSTM \\ Shared \\ Encoder \\ Decoder} & \shortstack{GRU \\ Last \\ Sequence} & \shortstack{GRU \\ Full \\ Sequence} & \shortstack{MLP \\ 1 \\ Layer \\ linear} & \shortstack{MLP \\ 2 \\ Layer \\ linear} & \shortstack{MLP \\ 3 \\ Layer \\ ReLU} & LinearNet & \shortstack{LinearNet \\ no \\ conv} & CLinear & NLinear & DLinear & \shortstack{TSMixer \\ 1} & \shortstack{TSMixer \\ 2} & \shortstack{TSMixer \\ no \\ relu \\ 1} & \shortstack{TSMixer \\ no \\ relu \\ 2} \\
\shortstack{n \\ ahead} & \shortstack{sequence \\ length} &  &  &  &  &  &  &  &  &  &  &  &  &  &  &  &  &  &  \\
\midrule
\multirow[t]{7}{*}{1} & 3 & 28 & 28 & 32251 & 694 & 1301 & 1321 & 28 & 581 & 1001 & 1066 & 170 & 14 & 14 & 18 & 202 & 372 & 372 & 372 \\
 & 10 & 91 & 91 & 32251 & 694 & 1301 & 1391 & 91 & 1841 & 2261 & 1108 & 184 & 21 & 21 & 32 & 433 & 827 & 827 & 827 \\
 & 20 & 181 & 181 & 32251 & 694 & 1301 & 1491 & 181 & 3641 & 4061 & 1168 & 204 & 31 & 31 & 52 & 933 & 1817 & 1817 & 1817 \\
 & 50 & 451 & 451 & 32251 & 694 & 1301 & 1791 & 451 & 9041 & 9461 & 1348 & 264 & 61 & 61 & 112 & 3633 & 7187 & 7187 & 7187 \\
 & 100 & 901 & 901 & 32251 & 694 & 1301 & 2291 & 901 & 18041 & 18461 & 1648 & 364 & 111 & 111 & 212 & 12133 & 24137 & 24137 & 24137 \\
 & 200 & nan & nan & 32251 & 694 & 1301 & 3291 & 1801 & 36041 & 36461 & 2248 & 564 & 211 & 211 & 412 & 44133 & 88037 & 88037 & 88037 \\
\bottomrule
\end{tabular}
\end{table}

\section{BTC volumes: weights}
\begin{figure}[H]
    \centering
    \includegraphics[width=1\textwidth]{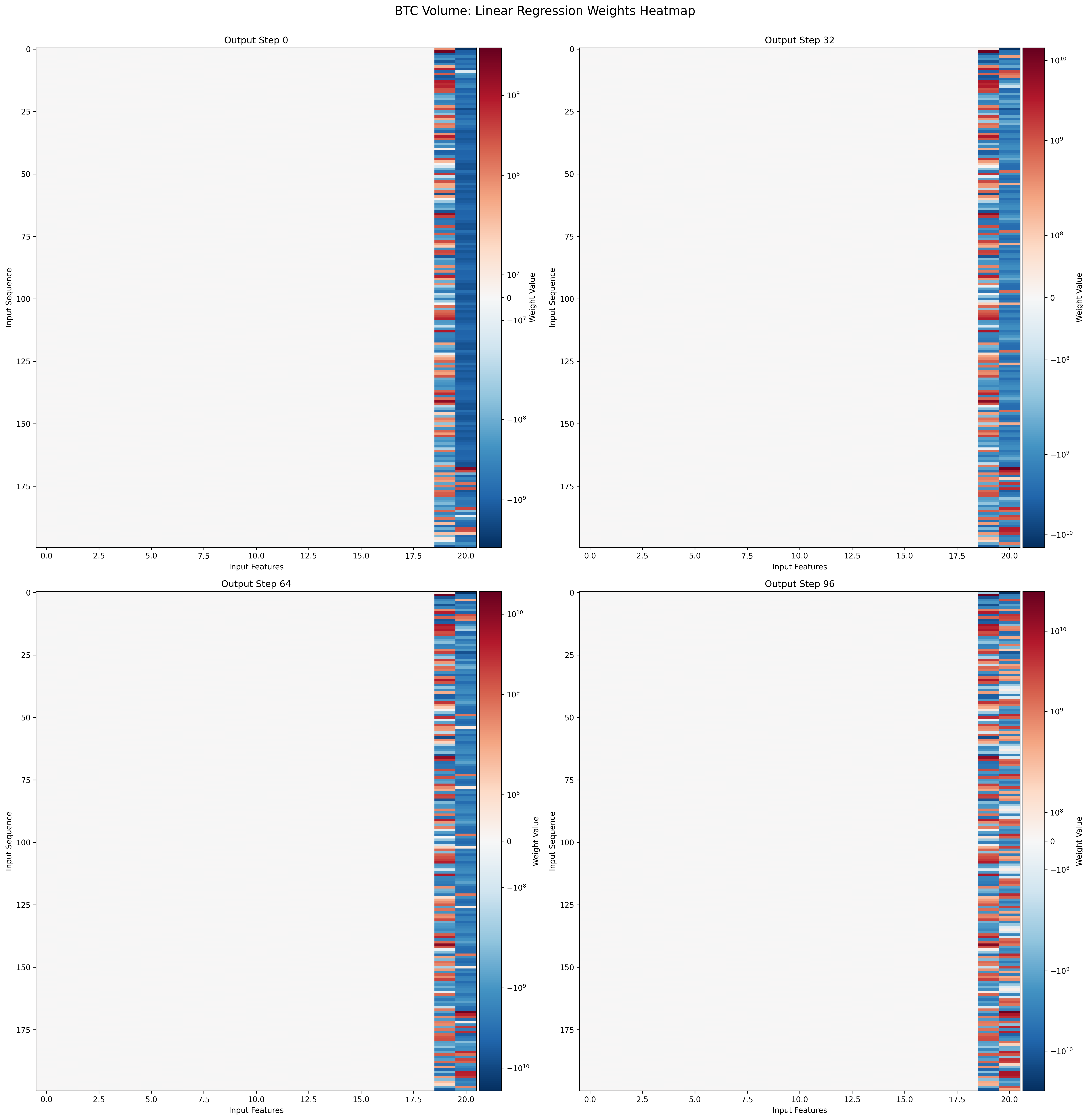}
    \caption{Linear Regression weights for BTC Volume task}
    \label{fig:etth1_linreg_weights}
\end{figure}

\begin{figure}[H]
    \centering
    \includegraphics[width=1\textwidth]{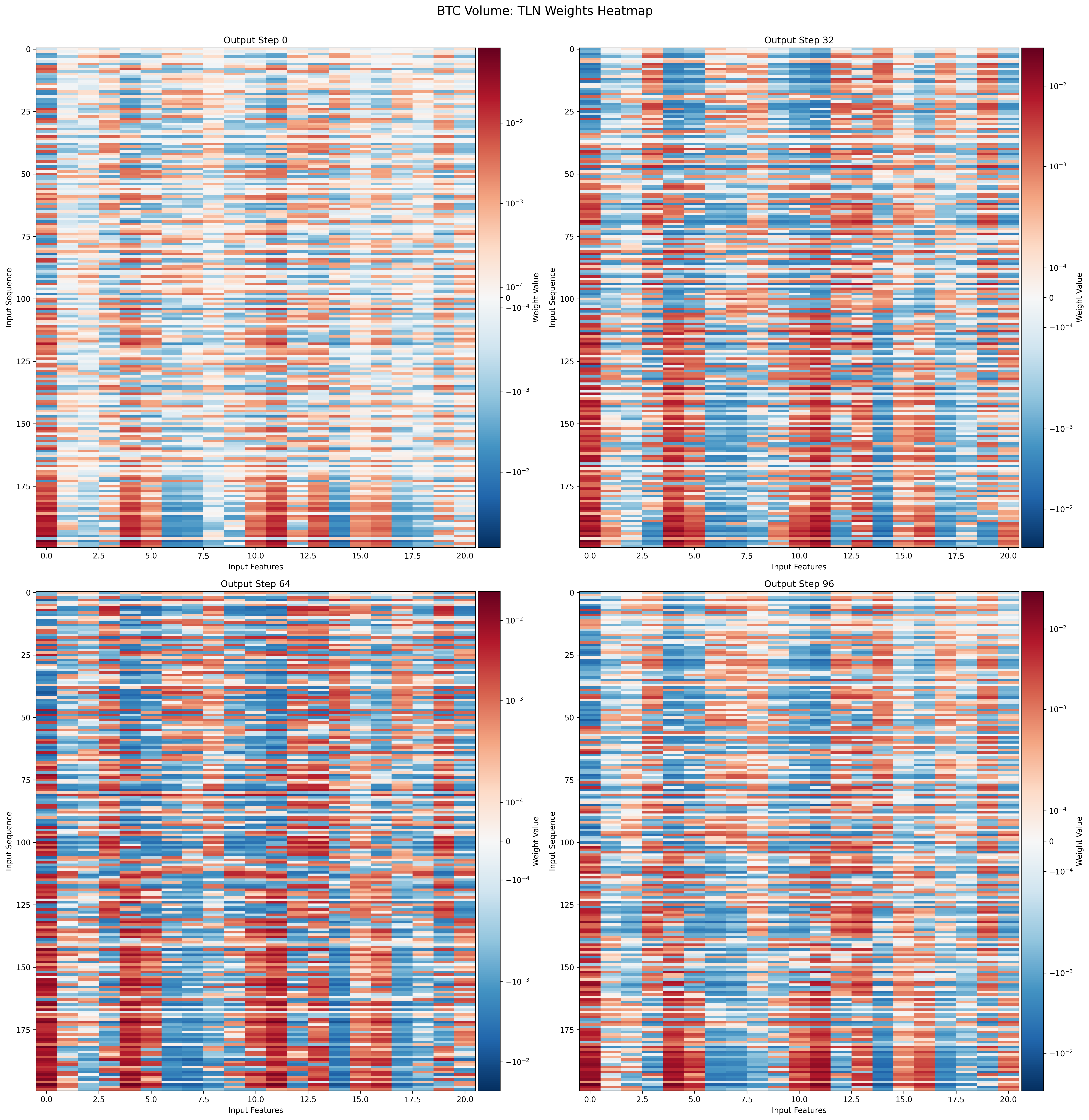}
    \caption{TLN weights for BTC Volume task}
    \label{fig:etth1_tln_weights}
\end{figure}

\begin{figure}[H]
    \centering
    \includegraphics[width=1\textwidth]{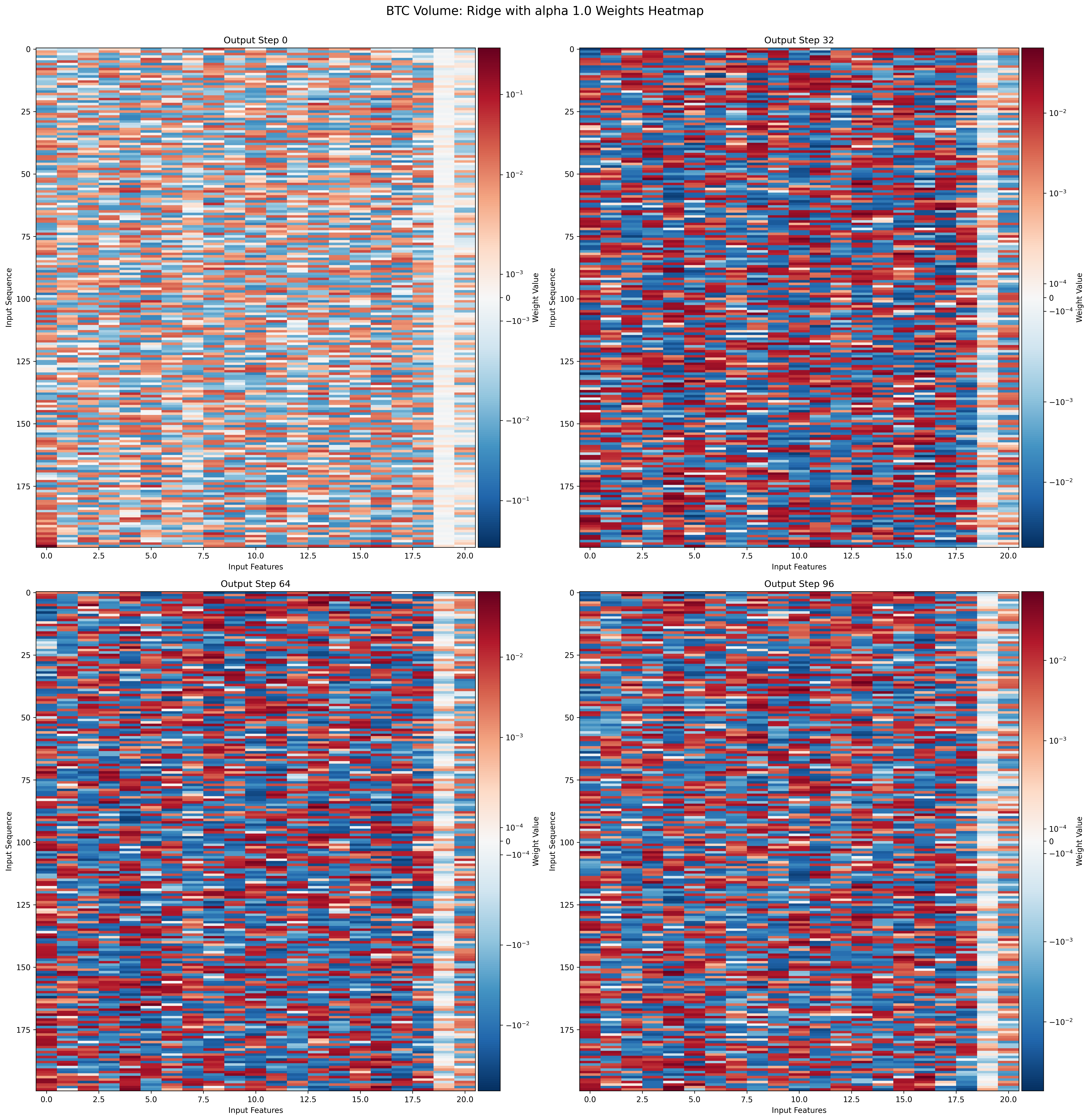}
    \caption{Ridge Regression weights for BTC Volume task (alpha = 1.0)}
    \label{fig:etth1_ridge_weights}
\end{figure}

\begin{figure}[H]
    \centering
    \includegraphics[width=1\textwidth]{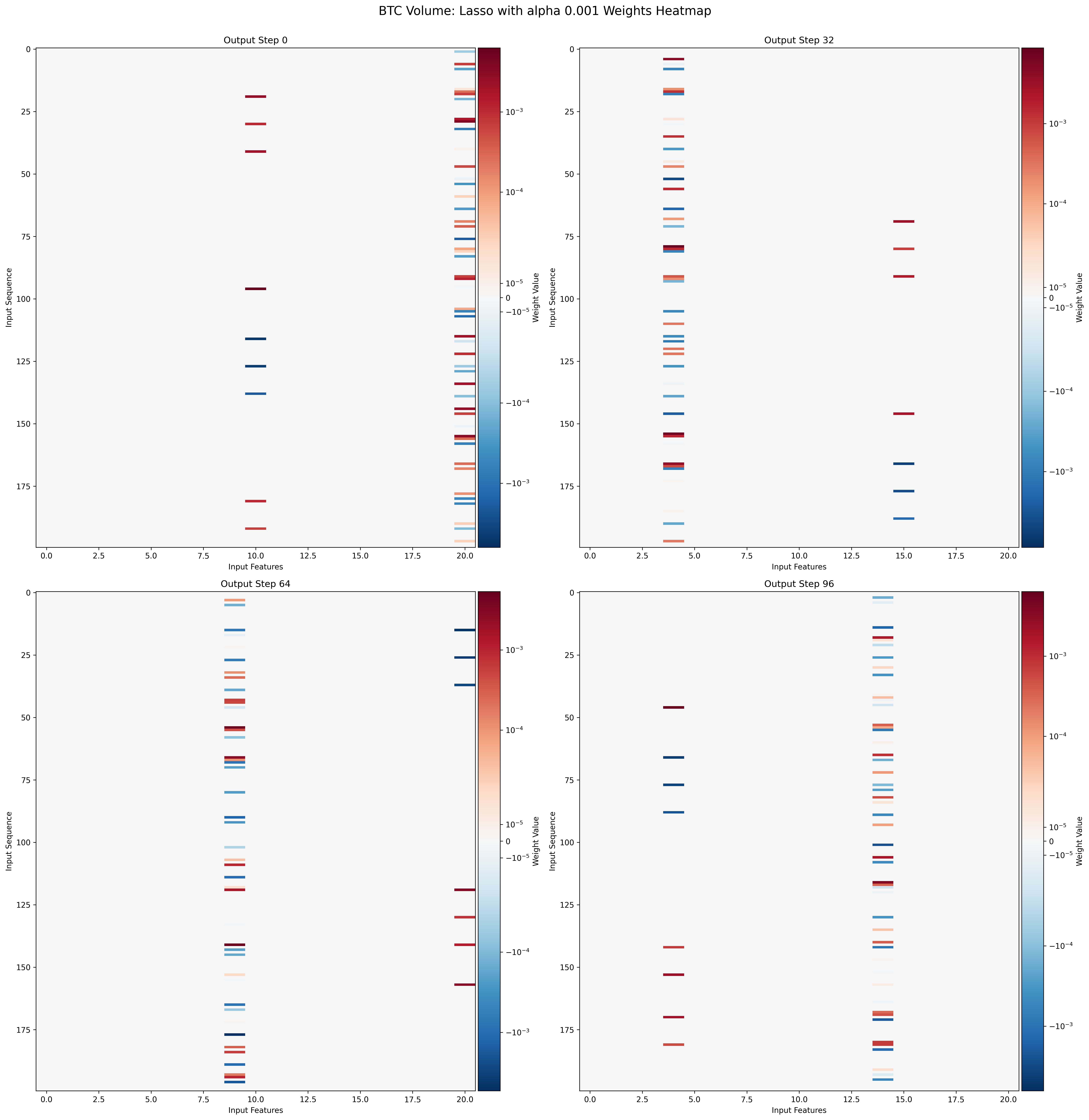}
    \caption{Lasso Regression weights for BTC Volume task (alpha = 0.01)}
    \label{fig:etth1_lasso_weights}
\end{figure}

\begin{figure}[H]
    \centering
    \includegraphics[width=1\textwidth]{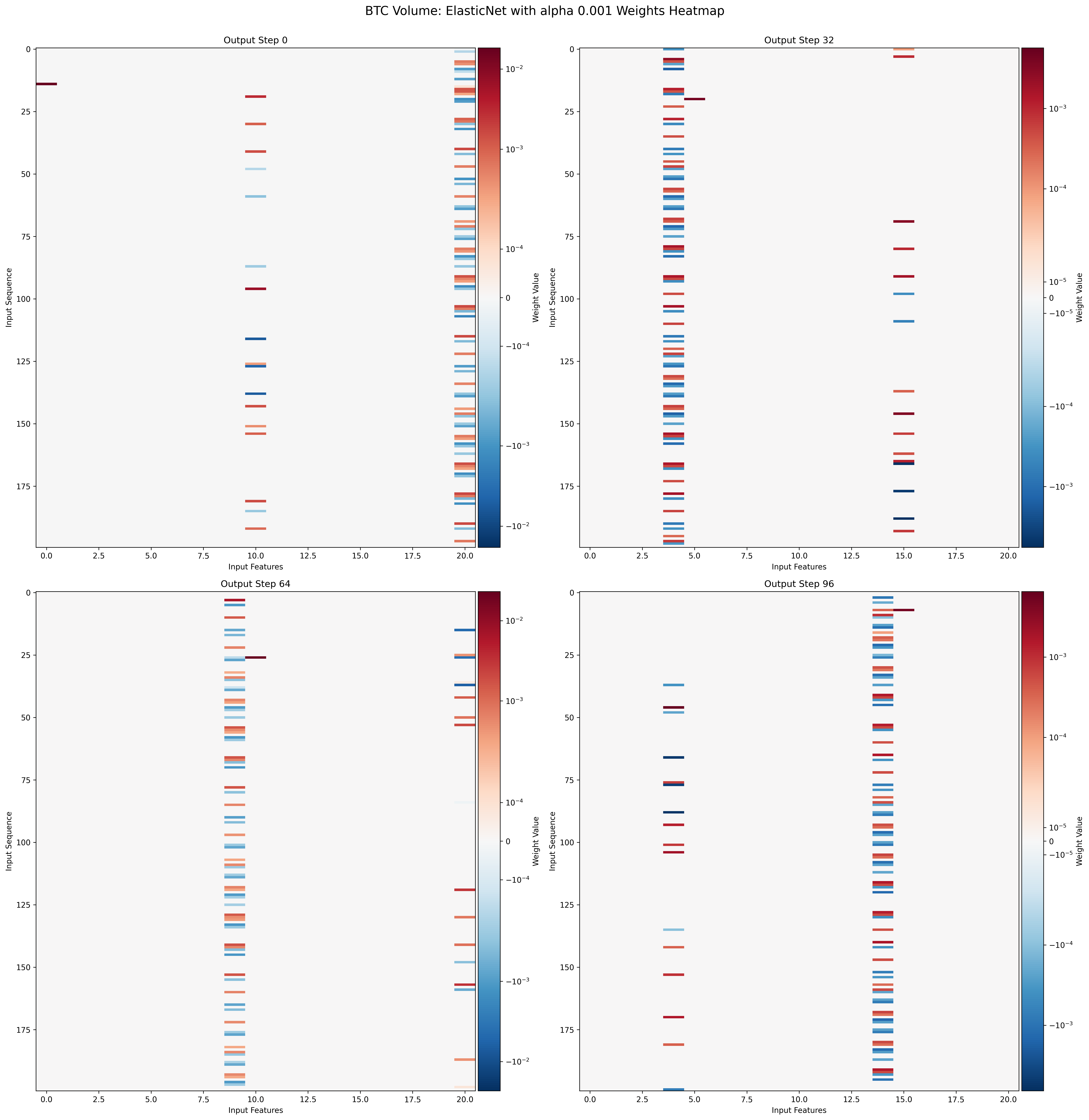}
    \caption{ElasticNet Regression weights for BTC Volume task (alpha = 0.01)}
    \label{fig:etth1_elastic_weights}
\end{figure}

\end{document}